\documentclass[11pt]{article}
\usepackage{listings}

\usepackage[final]{acl}

\usepackage{times}
\usepackage{latexsym}

\usepackage[T1]{fontenc}
\usepackage[utf8]{inputenc}

\usepackage{microtype}

\usepackage{inconsolata}

\usepackage{graphicx}

\usepackage{subcaption}
\usepackage{multirow}
\usepackage{tabularx,booktabs,array,seqsplit}
\newcolumntype{L}{>{\raggedright\arraybackslash}X}
\newcolumntype{C}{>{\centering\arraybackslash}m{1.6em}}
\usepackage{pifont}
\usepackage[dvipsnames]{xcolor}
\newcommand{\cmark}{\textcolor{ForestGreen}{\ding{51}}} % ✓
\newcommand{\xmark}{\textcolor{BrickRed}{\ding{55}}}    % ✗
\newcolumntype{L}{>{\raggedright\arraybackslash}X}
\newcolumntype{Y}{>{\centering\arraybackslash}p{1.50cm}}
\newcolumntype{S}{>{\raggedright\arraybackslash}p{0.32\textwidth}} % Name 고정폭
\newcolumntype{D}{>{\raggedright\arraybackslash}X}                           % Definition 가변
\newcolumntype{E}{>{\raggedright\arraybackslash}X}      

\title{PILOT-Bench: A Benchmark for Legal Reasoning in the Patent Domain with IRAC-Aligned Classification Tasks}

\author{
\textbf{Yehoon Jang}$^{1}$\thanks{\ \ Equal contribution.} \quad
\textbf{Chaewon Lee}$^{1}$\footnotemark[1] \quad
\textbf{Hyun-seok Min}$^{2}$ \quad
\textbf{Sungchul Choi}$^{1}$\thanks{\ \ Corresponding author.} \\
$^{1}$Major in Industrial Data Science \& Engineering,\\
Department of Industrial and Data Engineering, Pukyong National University \\
$^{2}$Tomocube Inc. \\
\texttt{\{jangyh0420, oochaewon\}@pukyong.ac.kr}, \texttt{min6284@gmail.com}, \texttt{sc82.choi@pknu.ac.kr}
}

\begin{document}
\maketitle
\begin{abstract}
The Patent Trial and Appeal Board (PTAB) of the USPTO adjudicates thousands of \textit{ex parte} appeals each year, requiring the integration of technical understanding and legal reasoning. While large language models (LLMs) are increasingly applied in patent and legal practice, their use has remained limited to lightweight tasks, with no established means of systematically evaluating their capacity for structured legal reasoning in the patent domain. To address this gap, we introduce \textbf{PILOT-Bench} (\textbf{P}atent \textbf{I}nva\textbf{L}idati\textbf{O}n \textbf{T}rial Benchmark), a dataset and benchmark that aligns PTAB decisions with USPTO patent data at the case-level and formalizes three IRAC-aligned classification tasks: Issue Type, Board Authorities, and Subdecision. We evaluate a diverse set of close-source(commercial) and open-source LLMs and conduct analyses across multiple perspectives, including input-variation settings, model families, and error tendencies. Notably, on the Issue Type task, closed-source(commercial) models consistently exceed 0.75 in Micro-F1 score, whereas the strongest open-source model (Qwen-8B) achieves performance around 0.56, highlighting the substantial gap in reasoning capabilities. PILOT-Bench establishes a foundation for the systematic evaluation of patent-domain legal reasoning and points toward future directions for improving LLMs through dataset design and model alignment. All data, code, and benchmark resources are available at \url{https://github.com/TeamLab/pilot-bench}.
\end{abstract}

\section{Introduction}

As the volume of patent applications and examinations continues to grow, the Patent Trial and Appeal Board (PTAB) of the US Patent and Trademark Office (USPTO) handles a substantial number of appeals and invalidation proceedings each year \cite{USPTO_PTAB_Statistics}. The \textit{ex parte} appeal, which challenges the rejection of an examiner, requires a precise interpretation of patent---such as claims and prior art---and legal reasoning to identify and apply the relevant provisions of 35 U.S.C. and 37 C.F.R. to reach a conclusion.

Large language models (LLMs) are increasingly used in patent and legal practice to reduce repetitive reading tasks \cite{AIGuidanceFR, AIPatentProcess, AutoPatent, USPTOAI}. However, their adoption remains largely limited to such lightweight tasks, while \textit{ex parte} appeals demand deep reasoning---issue identification, rule mapping, rule application, and conclusion determination---that go well beyond them. Furthermore, the lack of a systematic public dataset or benchmark hinders quantitative assessment of whether LLMs possess the technical understanding and legal reasoning required in PTAB invalidity review. As a result, using LLMs for these tasks remains challenging.

In this paper, we propose the \textbf{P}atent \textbf{I}nva\textbf{L}idati\textbf{O}n \textbf{T}rial Benchmark (PILOT-Bench), a dataset and benchmark for evaluating the legal reasoning abilities of LLMs in the patent domain. We combine PTAB decisions with USPTO data per case and construct classification tasks aligned with the Issue–Rule–Application–Conclusion (IRAC) framework commonly used in legal practice. Our contributions are threefold:
\begin{itemize}
\item \textbf{PILOT-Bench dataset \& benchmark.} PILOT-Bench is, to our knowledge, the first \emph{benchmark} that integrates 18K PTAB \textit{ex parte} appeals with USPTO patent text at the case-level and provides 15K opinion-split instances explicitly engineered to prevent label leakage.
\item \textbf{IRAC-aligned tasks.} We design three classification tasks; Issue Type(5 labels, multi-label), Board Authorities(9 labels, multi-label), Subdecision(23 fine/6 coarse grained labels, multi-class), directly aligned with the IRAC framework to measure patent-domain legal reasoning.
\item \textbf{Empirical evaluation.} We conduct input variation experiments to assess the respective contributions of role segmentation and claim-text augmentation across multiple LLMs.
\end{itemize}

PILOT-Bench establishes a benchmark for evaluating LLMs’ legal reasoning in the patent domain—specifically, PTAB \textit{ex parte} appeals where technical understanding and legal reasoning meet. Our objective is to open a durable, reusable point of comparison that can anchor subsequent model, data, and methodology work and, ultimately, support responsible use of LLMs in patent practice. Accordingly, we fix the evidence boundary via the Opinion Split: inputs contain only \texttt{appellant\_arguments} and \texttt{examiner\_findings}, with all \texttt{ptab\_opinion} text excluded. We keep the label schema fixed across Issue Type, Board Authorities, and Subdecision (fine/coarse) and evaluate under a unified zero-shot protocol with task-appropriate metrics (Exact Match/Macro-F1/Micro-F1 for multi-label; Accuracy/Macro-F1/Weighted-F1 for multi-class). We also report results for both closed-source(commercial) and open-source model families and for the Split (Base), Merge, and Split+Claim input-variation settings, providing reference baselines for subsequent work.

\section{Preliminaries}

\subsection{PTAB \textit{ex parte} Appeal}
The PTAB \textit{ex parte} appeal process is initiated after a final rejection by a patent examiner. The appellant submits an Appeal Brief, followed by an Examiner’s Answer and, optionally, a Reply Brief. The Board then issues a written decision. PTAB decisions are conventionally organized into sections such as the \textit{Statement of the Case}, outlining the procedural and factual background, and the \textit{Analysis}, presenting the legal reasoning. The concluding portion records the outcome at the claim or case-level and cites the statutory or regulatory authorities (e.g., 35 U.S.C., 37 C.F.R.) that ground the ruling. In this way, PTAB decisions closely reflect the flow of legal reasoning.

\subsection{IRAC Framework}
% The IRAC framework is a widely adopted method for structuring legal analysis \citep{IRAC}.  In this framework, the \textit{Issue} refers to the identification of the legal questions raised by the facts of the case. The \textit{Rule} stage maps the identified issues to their governing legal provisions. The \textit{Application} stage involves applying the identified rules to the specific facts and arguments at issue.  Finally, the \textit{Conclusion} states the ultimate legal outcome derived from this reasoning process.
% In the context of PTAB \textit{ex parte} appeals, the IRAC stages map directly onto the PTAB decision process. The Issue step identifies contested statutory requirements (e.g., novelty, non-obviousness). The Rule step involves citing the relevant sections of U.S. patent law and procedural regulations. The Application step weighs the appellant’s and examiner’s arguments against prior art and statutory requirements. Finally, the Conclusion step determines the Board’s ruling at the claim level.
In PTAB \textit{ex parte} appeals, IRAC maps naturally onto the decision flow: Issue identifies the contested statutory grounds; Rule maps those issues to the governing legal provisions; Application weighs the parties’ arguments and facts against those provisions; and Conclusion renders the Board’s ruling. We operationalize Issue, Rule, and Conclusion as three classification tasks and leave Application to future, generation-based work.

Our benchmark translates three of these IRAC stages---Issue, Rule, and Conclusion---into three concrete classification tasks to evaluate LLMs’ capacity for patent-domain legal reasoning.

\begin{table}[t]
  \centering
  \small
  \renewcommand{\arraystretch}{0.98}
  \setlength{\tabcolsep}{6pt}
  \begin{tabularx}{\columnwidth}{L C C C}
    \toprule
    \textbf{Dataset / Study} & \textbf{Patent} & \textbf{Legal} & \textbf{LLM} \\
    \midrule
    \multicolumn{4}{l}{\textbf{Patent}} \\
    \midrule
    WIPO-alpha      & \cmark & \xmark & \xmark \\
    CLEF-IP         & \cmark & \xmark & \xmark \\
    USPTO-2M        & \cmark & \xmark & \xmark \\
    BIGPATENT       & \cmark & \xmark & \xmark \\
    HUPD            & \cmark & \xmark & \cmark \\
    IMPACT          & \cmark & \xmark & \cmark \\
    Patent-CR       & \cmark & \xmark & \cmark \\
    \midrule
    \multicolumn{4}{l}{\textbf{Legal}} \\
    \midrule
    LegalBench      & \xmark & \cmark & \cmark \\
    LexGLUE         & \xmark & \cmark & \xmark \\
    CaseHOLD        & \xmark & \cmark & \xmark \\
    CUAD / LEDGAR\footnotemark & \xmark & \xmark & \xmark \\
    Pile of Law     & \xmark & \cmark & \xmark \\
    MultiLegalPile  & \xmark & \cmark & \xmark \\
    \midrule
    \multicolumn{4}{l}{\textbf{PTAB studies}} \\
    \midrule
    Winer (2017)               & \cmark & \cmark & \xmark \\
    Rajshekhar (2017)          & \cmark & \xmark & \xmark \\
    Love (2019)                & \cmark & \cmark & \xmark \\
    Garcia (2022)              & \cmark & \cmark & \xmark \\
    Sokhansanj \& Rosen (2022) & \cmark & \cmark & \xmark \\
    Fu (2021)                  & \cmark & \xmark & \xmark \\
    \midrule
    \textbf{PILOT-Bench} & \textbf{\cmark} & \textbf{\cmark} & \textbf{\cmark} \\
    \bottomrule
  \end{tabularx}
  \caption{Comparison by three criteria: (1) patent tasks, (2) legal/adjudicatory tasks, (3) ability to evaluate LLM in the patent/legal domain. Legal/adjudicatory tasks denote tasks leveraging statutory/regulatory mappings and decision structure. PTAB entries are research studies (not reusable corpora).}
  \label{tab:dataset-comparison}
\end{table}

\footnotetext{CUAD/LEDGAR focus on contract clause extraction/classification; they are not decision/holding–centric and do not map statutes/regulations, hence marked \xmark\ under Legal/adjudicatory.}

\begin{figure*}[t]
    \includegraphics[width=\linewidth]{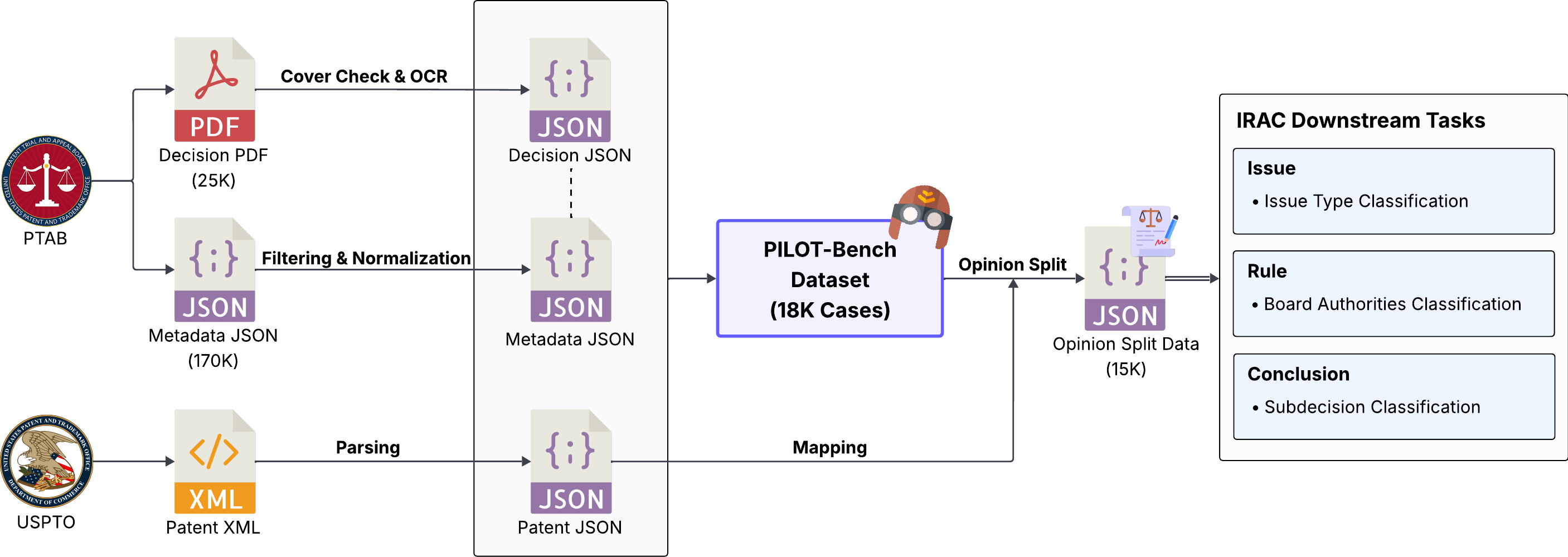}
    \caption{PILOT-Bench: Data sources, processing pipeline, and tasks. PTAB metadata JSONs and decision JSONs are aligned with USPTO patent JSONs to form PILOT-Bench (18K). From this base, we map each case to the appellant’s patent and apply an LLM opinion split, yielding the 15K Opinion Split Data used for IRAC-aligned classification tasks.}
    \label{fig:overview}
\end{figure*}

\section{Related Work}

\subsection{Patent Corpora/Benchmarks}
Public patent corpora have largely been constructed around technical-text tasks such as summarization and classification. WIPO-alpha \cite{Wipo-2M}, CLEF-IP \cite{2010CLEFIP, 2011CLEFIP}, and USPTO-2M \cite{DeepPatent} provide patent full text together with bibliographic metadata and introduce evaluation setups for IPC/CPC classification and prior-art retrieval research. BIGPATENT \cite{BIGPATENT} releases roughly 1.3 million description–abstract pairs and establishes a long-document summarization benchmark. HUPD \cite{HUPD} links patent documents filings from 2004–2018 with metadata, enabling multiple tasks including classification and binary decision prediction. More recently, IMPACT \cite{IMPACT} introduces a multimodal dataset by combining design images with patent information, while Patent-CR \cite{PatentCR} expands the scope of patent datasets by defining a claim-centric corpus for claim-revision tasks.  

\subsection{Legal Corpora/Benchmarks}
LegalBench \cite{LegalBench} covers legal reasoning broadly with 162 tasks and defines IRAC-stage tasks. LexGLUE \cite{lexglue} is a multi-task legal NLU benchmark that offers evaluation setups for case classification, topic classification, and clause identification in contracts. CUAD \cite{CUAD} and LEDGAR \cite{ledgar} construct clause extraction and classification tasks from contracts. CaseHOLD \cite{CaseHOLD} targets holding identification within judicial opinions. Pile of Law \cite{PileOfLaw} and MultiLegalPile \cite{multilegalpile} offer large-scale pretraining corpora aggregating diverse legal subdomains.  

\subsection{PTAB Studies}
Prior PTAB prediction and analysis studies can be organized by procedure type and input modality. \citet{Winer2017UCBEECS} targets Post-Grant Review (PGR) disputes and uses SVM and random forests to predict institution and invalidation outcomes. \citet{Rajshekhar2017ICAIL} works in \textit{Ex Parte} Reexamination (EPR), performing prior-art retrieval from the abstract, the first claim, and the title. \citet{Love2019Determinants} studies Inter Partes Review (IPR), predicting institution from metadata such as the number of unique words in the first independent claim and specification length. \citet{Garcia2022RAIL} combines claims with rejection grounds and classifies PTAB final decisions using BERT. \citet{SokhansanjRosen2022ApplSci} uses the Patent Owner Preliminary Response (POPR) and decision text as inputs and applies XGBoost and a CNN-Attention model to predict IPR institution. \citet{AbilityOrChoice} leverages IPR institution and final outcomes to estimate firm-level patent performance measures.  

\paragraph{Limitations across Domains.}
Taken together, these studies reveal persistent gaps across patent, legal, and PTAB corpora. Patent benchmarks remain confined to technical-text problems such as summarization, classification, and retrieval, without capturing legal reasoning grounded in statutory authorities or decision structure. Legal corpora address reasoning tasks broadly, yet largely overlook the patent domain. PTAB studies have primarily examined procedures distinct from \textit{ex parte} appeal, such as Post-Grant Review (PGR), Inter Partes Review (IPR), and \textit{Ex Parte} Reexamination (EPR), or have focused on predicting outcomes from text and metadata, with little attention to integrated legal reasoning or LLM evaluation.  

\textbf{PILOT-Bench} directly addresses these shortcomings by targeting \textit{ex parte} appeals, aligning PTAB decisions with USPTO patent information at the case-level, and translating the IRAC framework into classification tasks that enable systematic assessment of LLMs’ legal-reasoning performance in the patent domain.

\section{Data Construction}

This section describes the construction of the PILOT-Bench dataset, including source collection, case-level alignment, text normalization, opinion splitting, and label refinement. The goals are threefold: (i) to consistently align PTAB decisions with USPTO patent information; (ii) to prevent answer leakage by excluding the Board’s opinion from inputs via the Opinion Split; and (iii) to provide input–label sets that reflect PTAB practice and are directly applicable to IRAC-aligned classification tasks.

\begin{figure}[t]
    \includegraphics[width=\linewidth]{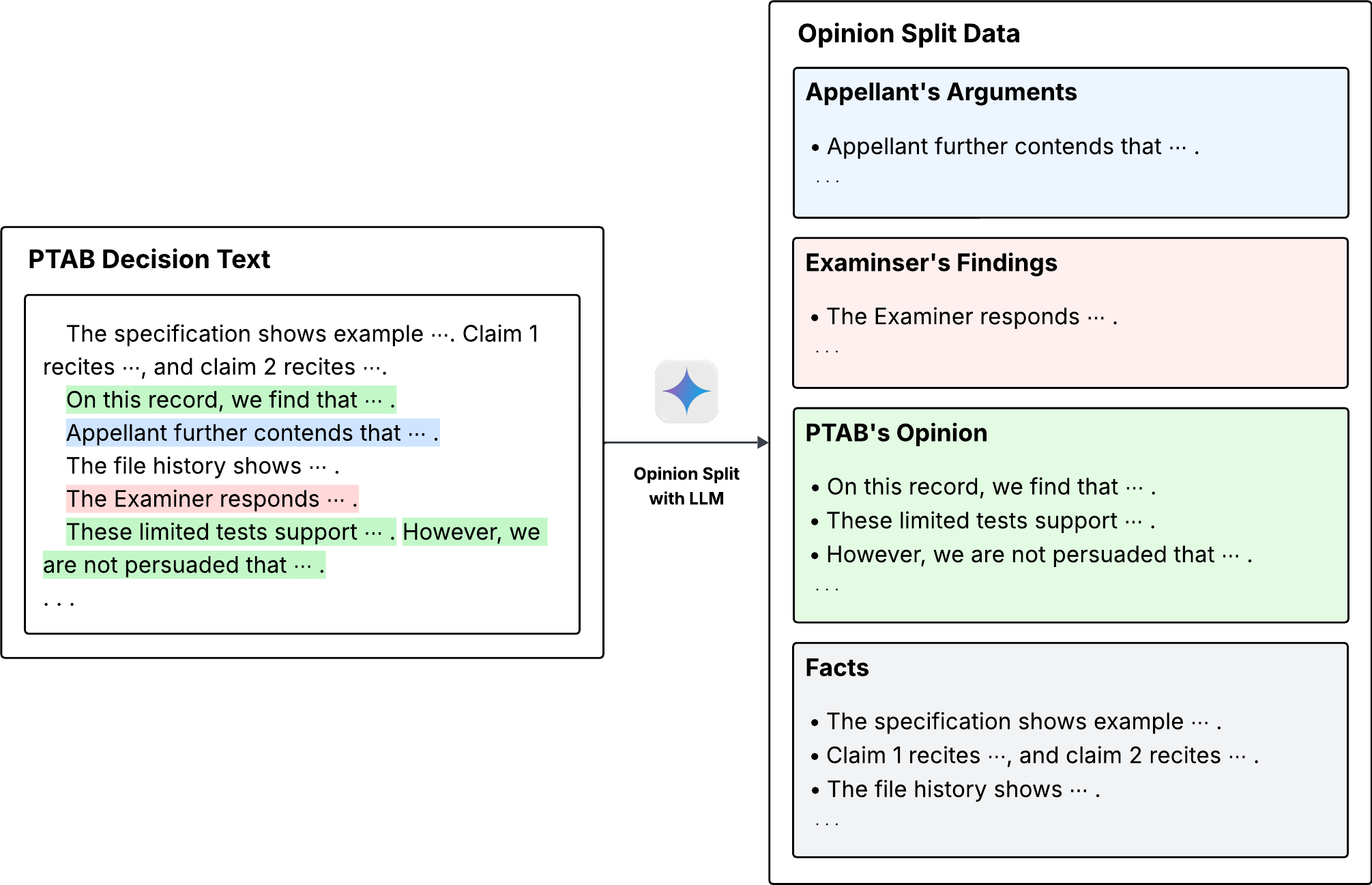}
    \caption{Opinion Split of PTAB Decisions. Given a PTAB decision, an LLM segments the text at the sentence-level and, using context, classifies each sentence into four roles; \texttt{appellant\_arguments}, \texttt{examiner\_findings}, \texttt{ptab\_opinion}, and \texttt{facts}. The resulting Opinion Split Data serves as the base input for our IRAC-aligned classification tasks.}
    \label{fig:opinion_split}
\end{figure}

\subsection{Data Sources \& Scope}
\begin{itemize}
    \item \textbf{PTAB Metadata (JSON, 170K)} Using USPTO’s PTAB API v2\footnote{\url{https://developer.uspto.gov/api-catalog/ptab-api-v2}}, we collect metadata such as proceeding identifiers, application/publication numbers, proceeding type, panel judges, decision dates, and decision types.
    \item \textbf{PTAB Decisions (PDF, 25K)} We apply OCR to the original PDF decisions to extract the full opinion text and segment conventional sections such as \textit{Decision on Appeal}, \textit{Statement of the Case}, and \textit{Analysis}.
    \item \textbf{USPTO Patent (XML)} From USPTO bulk XML\footnote{\url{https://data.uspto.gov/bulkdata/datasets}}, we extract only textual components---titles, claims, and specifications---and preprocess claims to preserve their dependency structures.
\end{itemize}

We set the PTAB window to 2009–2024 to ensure consistent document formatting and reliable OCR (standardized cover pages). For USPTO patent text, we use 2006–2024 to approximate a 20-year horizon relative to appeal filings and to cover applications linked to appeals decided after 2009.

\subsection{Opinion Split}

PTAB decisions intermix the appellant’s arguments, the examiner’s findings, and the PTAB’s opinion. To prevent answer leakage, we exclude the Board’s opinion from model inputs and retain only the appellant’s and examiner’s arguments. This design ensures that classification tasks such as Issue Type, Board Authorities, and Subdecision measure an LLM’s ability to compare and synthesize conflicting arguments, rather than relying on the Board’s conclusions.

The split dataset is primarily derived from the \textit{Statement of the Case} and \textit{Analysis} sections, which encompass the substantive exchanges between the appellant and the examiner. To construct the split dataset, each decision is processed by an LLM instructed to classify sentences into four categories: \texttt{appellant\_arguments}, \texttt{examiner\_findings}, \texttt{ptab\_opinion}, and \texttt{facts}. After evaluating outputs across multiple models, we selected Gemini-2.5-pro as the final splitter for large-scale classification. The full prompt used in this task is provided in the Appendix \ref{sec:task-prompt}.

In addition, we further analyzed document-level statistics of the Opinion Split data to assess input scale and variability across decisions. On average, each split decision contains approximately 1.4K words and 8.7K characters, reduced by about 25\% relative to the original sections (\textit{Statement of the Case} + \textit{Analysis}) due to the exclusion of PTAB opinion text. Among the original sections, the \textit{Statement of the Case} averages 430 words while the \textit{Analysis} section averages 1.4K words, indicating that most of the reasoning content resides in the latter. Within the split data, the \texttt{appellant\_arguments} and \texttt{examiner\_findings} segments are similar in length (about 300 words each), whereas the \texttt{ptab\_opinion} portion, retained only for reference, is substantially longer and more variable (820 words on average). These findings suggest that the input texts used for model evaluation maintain a balanced representation of opposing arguments while preserving realistic document scale. Full descriptive statistics, including word- and character-level summaries and role-wise distributions, are provided in Appendix~\ref{sec:appendix-length-stats}.

% \subsection{USPTO Patent Linkage across PTAB}

% To assess the structural linkage between PTAB cases and USPTO patents, we further examined the number of patents associated with each case after alignment. On average, a PTAB case links to approximately 2.05 patents (one appellant patent and about one prior patent), with a few complex cases referencing up to 14 distinct prior patents. This connectivity analysis confirms that most appeals involve one primary patent and a limited number of prior-art references, consistent with typical \textit{ex parte} appeal practice (see Appendix~\ref{sec:appendix-linked-case}).

\begin{figure*}[t]
  \centering
  \begin{subfigure}{0.24\textwidth}
    \centering
    \includegraphics[width=\linewidth]{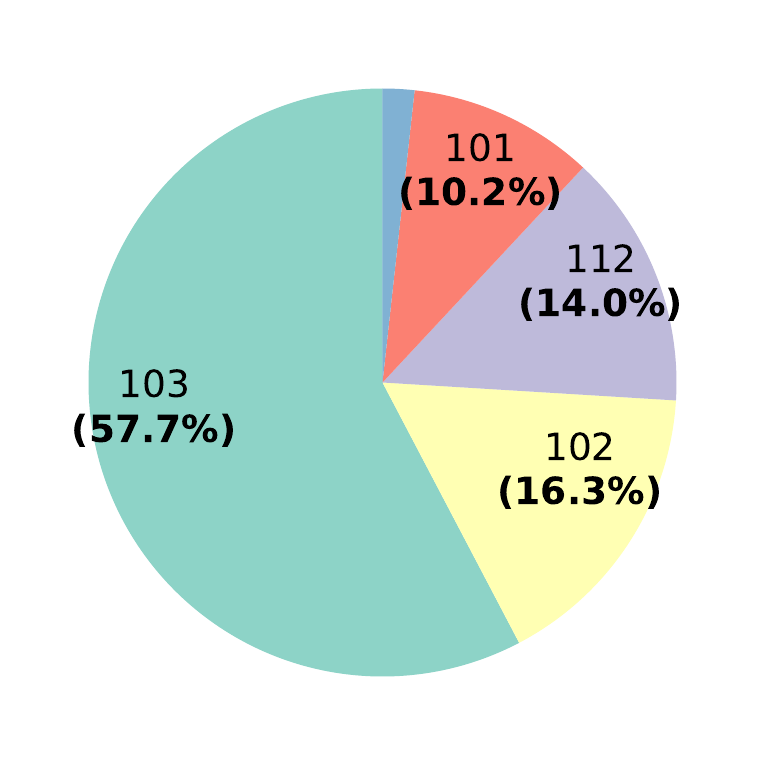}
    \caption{Issue type}
    \label{fig:dist-issue}
  \end{subfigure}\hfill
  \begin{subfigure}{0.24\textwidth}
    \centering
    \includegraphics[width=\linewidth]{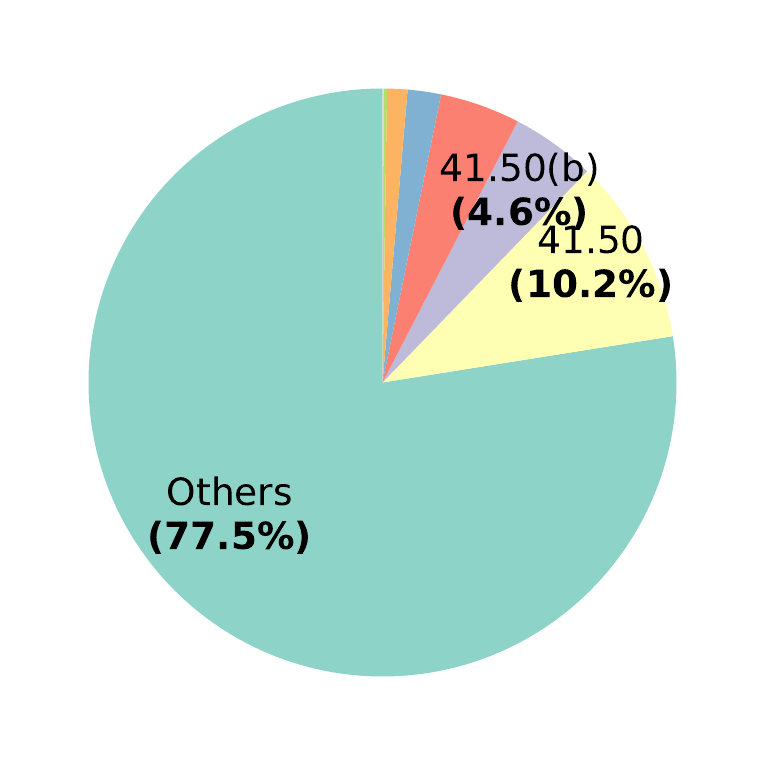}
    \caption{Board Authorities}
    \label{fig:dist-authorities}
  \end{subfigure}\hfill
  \begin{subfigure}{0.24\textwidth}
    \centering
    \includegraphics[width=\linewidth]{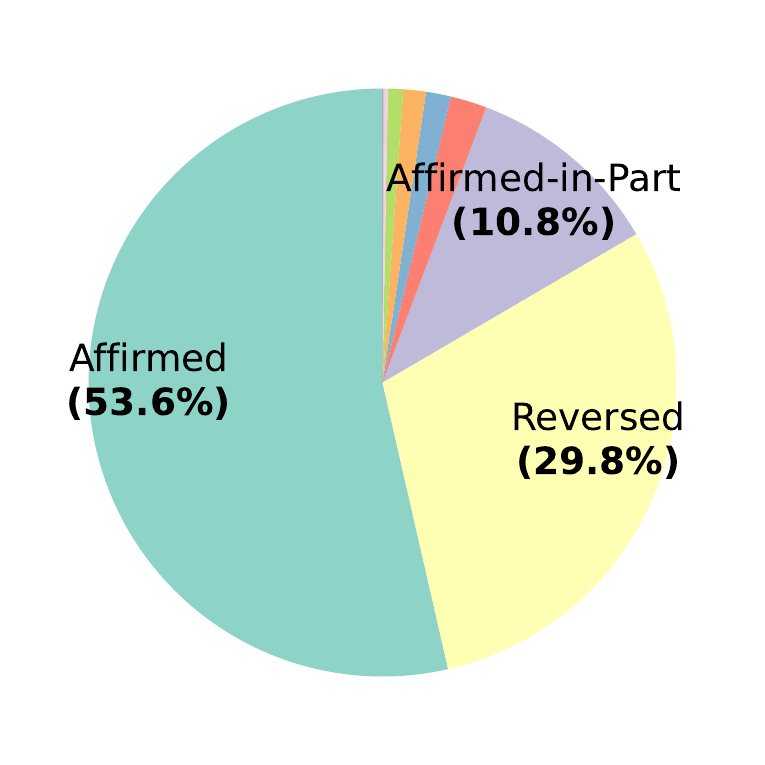}
    \caption{Subdecision (Fine, Top10)}
    \label{fig:dist-subdecision-fine}
  \end{subfigure}\hfill
  \begin{subfigure}{0.24\textwidth}
    \centering
    \includegraphics[width=\linewidth]{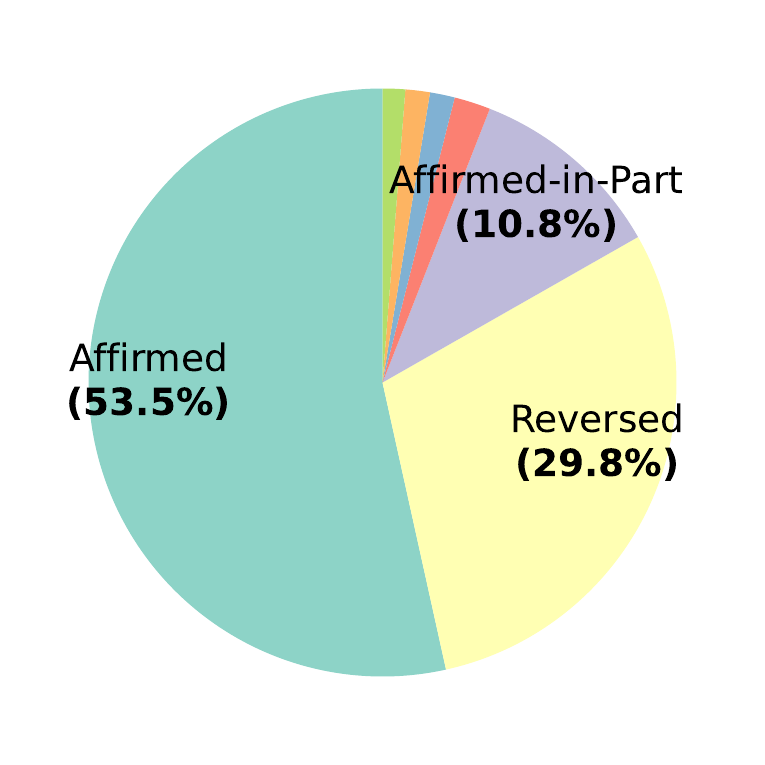}
    \caption{Subdecision (Coarse)}
    \label{fig:dist-subdecision-coarse}
  \end{subfigure}

  \caption{Label distributions across tasks are imbalanced; for Subdecision (fine), only the top 10 labels are shown. Bold values under the labels are the proportion each label occupies in the dataset.}
  \label{fig:label-dist}
\end{figure*}

\subsection{Labeling Sources \& Regularization}

We refine labels for three classification tasks, starting from the metadata in PTAB JSON and consolidating them into a schema restricted to merits determinations in \textit{ex parte} appeals.  

For the Issue Type task, the raw metadata contained six statutory sections under 35~U.S.C. (\S 100, 101, 102, 103, 112, and 120). To improve consistency and focus on the most frequent and practically relevant issues, we reduced these to five labels: \textit{101}, \textit{102}, \textit{103}, \textit{112}, and an \textit{Others} category. Because a single appeal may raise multiple issues, this task is modeled as multi-label.  

For the Board Authorities task, we identified the regulatory provisions cited in PTAB's opinions as the operative authorities for decisions. Although 35 U.S.C. sections appear in the raw data, the operative authority in \textit{ex parte} appeals is generally 37 C.F.R.; accordingly, we select the most frequent provisions---\S \textit{1.131}, \textit{1.132}, \textit{41.50}, \textit{41.50(a)}, \textit{41.50(b)}, \textit{41.50(c)}, \textit{41.50(d)}, and \textit{41.50(f)}---and group the remainder under \textit{Others}, yielding a nine-label schema. Boilerplate references such as 35~U.S.C. \S134 were excluded. Like Issue Type, this task is modeled as multi-label.  

For the Subdecision task, we standardized the final outcomes of PTAB decisions. In the base dataset, we initially observed 34 distinct outcome labels. Since our corpus is restricted to appeal proceedings, we excluded reexamination appeals as well as AIA trial outcomes (e.g., IPR, PGR, CBM), removing AIA-specific categories such as Institution Granted. This reduction yielded 23 appeal-specific outcomes. We then applied normalization (case folding, whitespace and punctuation unification) and synonym merging to consolidate the labels. We provide these 23 outcomes as a set of fine-grained labels, which include an \textit{Others} category grouping infrequent outcomes. In addition, we map them into six coarse-grained labels that dominate in \textit{ex parte} appeals: \textit{Affirmed}, \textit{Affirmed with New Ground of Rejection}, \textit{Affirmed-in-Part}, \textit{Affirmed-in-Part with New Ground of Rejection}, \textit{Reversed}, \textit{Reversed with New Ground of Rejection}, and \textit{Others}.  

After defining these schemas, we examined their distributions. As shown in Figure~\ref{fig:label-dist}, all tasks are highly imbalanced. Additional information on the labels is provided in the Appendix \ref{sec:label}.

\begin{figure}[t]
    \includegraphics[width=\linewidth]{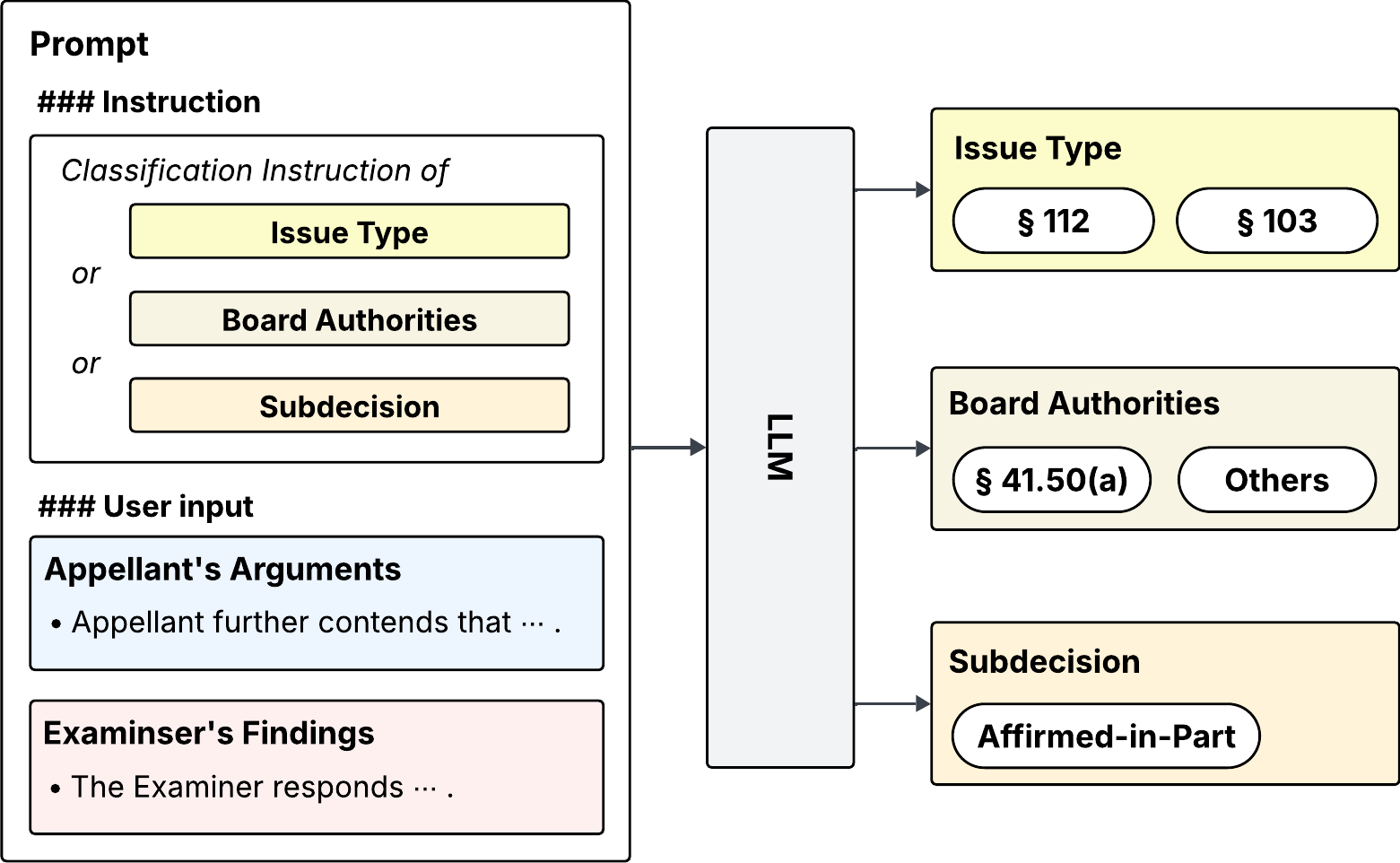}
    \caption{Task-specific prompting. A standardized prompt combines a task-specific instruction with the \texttt{appellant\_arguments} and \texttt{examiner\_findings} segments; the LLM then executes the chosen task--Issue, Board Authorities, or Subdecision--and outputs from the predefined label set.}
    \label{fig:task}
\end{figure}

\begin{table*}[t]
  \centering
  \scriptsize
  % ===== 왼쪽 표 =====
  \begin{subtable}{0.48\textwidth}
    \centering
    \begin{tabular}{lccc}
      \toprule
      \textbf{Model} & \textbf{Exact Match} & \textbf{Macro-F1} & \textbf{Micro-F1}\\
%  ===================== BASE block =====================
      \midrule
        \multicolumn{4}{c}{\textbf{Split (Base)}} \\
        \cmidrule(lr){1-4}
        Claude-Sonnet-4 & 0.5871	& 0.5457	& 0.7905  \\
        Gemini-2.5-pro  & 0.5874    & 0.6630	& 0.7923  \\
        GPT-4o          & 0.5751    & 0.6519    & 0.7860  \\
        GPT-o3          & \textbf{0.5955}	& \textbf{0.6639}	& \textbf{0.7968}  \\
        Solar-pro2      & 0.5583	& 0.5240	& 0.7707  \\
        \addlinespace[2pt]
        LLaMA-3.1(8B)       & 0.1826	& 0.1051	& 0.5793  \\
        Mistral(7B)         & 0.3405    & 0.2111	& 0.6080  \\
        Qwen(8B)            & 0.5561	& 0.5251	& 0.7741  \\
        T5(2B)              & 0.0772	& 0.3845	& 0.4469  \\
        \midrule
% ===================== MERGE block =====================
        \multicolumn{4}{c}{\textbf{Merge}} \\
        \cmidrule(lr){1-4}
        Claude-Sonnet-4 & 0.5879	& 0.5468    & 0.7915  \\
        Gemini-2.5-pro  & 0.5810    & 0.6625    & 0.7889  \\
        GPT-4o          & 0.5516	& 0.6422	& 0.7758  \\
        GPT-o3          & \textbf{0.5943}	& \textbf{0.6645}    & \textbf{0.7961}  \\
        Solar-pro2      & 0.5466	& 0.6249    & 0.7643  \\
        \addlinespace[2pt]
        LLaMA-3.1(8B)       & 0.1334	& 0.4517	& 0.5801  \\
        Mistral(7B)         & 0.2639	& 0.1356    & 0.5760  \\
        Qwen(8B)            & 0.5322    & 0.6255    & 0.7634  \\
        T5(2B)              & 0.0057    & 0.3534    & 0.4050  \\
        \midrule
% ===================== CLAIM block =====================
        \multicolumn{4}{c}{\textbf{Split+Claim}} \\
        \cmidrule(lr){1-4}
        Claude-Sonnet-4 & 0.5869    & 0.5443    & 0.7915  \\
        Gemini-2.5-pro  & 0.5911    & 0.6632    & 0.7955   \\
        GPT-4o          & 0.5658	& 0.6492    & 0.7828  \\
        GPT-o3          & \textbf{0.5946}	& \textbf{0.6639}	& \textbf{0.7967}  \\
        Solar-pro2      & 0.5355    & 0.6225	& 0.7596  \\
        \addlinespace[2pt]
        LLaMA-3.1(8B)       & 0.1785	& 0.4360  	& 0.5928  \\
        Mistral(7B)         & 0.4200	& 0.2662	& 0.6767  \\
        Qwen(8B)            & 0.5631    & 0.6353    & 0.7782  \\
        T5(2B)              & 0.0155    & 0.0024    & 0.4545  \\
    \bottomrule
    \end{tabular}
    \caption{Issue Type}
    \label{tab:issue-type_main_result}
  \end{subtable}\hfill
  % ===== 오른쪽 표 =====
  \begin{subtable}{0.48\textwidth}
    \centering
    \begin{tabular}{lccc}
      \toprule
      \textbf{Model} & \textbf{Exact Match} & \textbf{Macro-F1} & \textbf{Micro-F1}\\
% ===================== BASE block =====================
      \midrule
        \multicolumn{4}{c}{\textbf{Split (Base)}} \\
        \cmidrule(lr){1-4}
        Claude-Sonnet-4 & 0.4945	& 0.2397	& 0.5444  \\
        Gemini-2.5-pro  & 0.5906    & \textbf{0.2665}    & \textbf{0.6916}  \\
        GPT-4o          & \textbf{0.6314}    & 0.2589	& 0.6522  \\
        GPT-o3          & 0.5302	& 0.1940    & 0.6236  \\
        Solar-pro2      & 0.4293	& 0.1014	& 0.6179  \\
        \addlinespace[2pt]
        LLaMA-3.1(8B)       & 0.0000    & 0.0843    & 0.1230  \\
        Mistral(7B)         & 0.0028    & 0.0075    & 0.2762  \\
        Qwen(8B)            & 0.1542    & 0.1420    & 0.1966  \\
        T5(2B)              & 0.0064    & 0.0026    & 0.2116  \\
        \midrule
% ===================== MERGE block =====================
        \multicolumn{4}{c}{\textbf{Merge}} \\
        \cmidrule(lr){1-4}
        Claude-Sonnet-4 & \textbf{0.7761}    & 0.2128    & \textbf{0.8033}  \\
        Gemini-2.5-pro  & 0.6323    & 0.3062    & 0.7387  \\
        GPT-4o          & 0.6032    & \textbf{0.2486}    & 0.6179  \\
        GPT-o3          & 0.6459    & 0.2160	& 0.7344  \\
        Solar-pro2      & 0.2531    & 0.0620	& 0.5524  \\
        \addlinespace[2pt]
        LLaMA-3.1(8B)       & 0.0000    & 0.0882	& 0.1629  \\
        Mistral(7B)         & 0.0028	& 0.0038    & 0.2729  \\
        Qwen(8B)            & 0.4266    & 0.1897    & 0.4531  \\
        T5(2B)              & 0.0026	& 0.0032    & 0.1757  \\
        \midrule
% ===================== CLAIM block =====================
        \multicolumn{4}{c}{\textbf{Split+Claim}} \\
        \cmidrule(lr){1-4}
        Claude-Sonnet-4 & 0.2026    & 0.1530    & 0.2636  \\
        Gemini-2.5-pro  & \textbf{0.4913}    & \textbf{0.2201}    & \textbf{0.5795}  \\
        GPT-4o          & 0.0035    & 0.1425    & 0.1431  \\
        GPT-o3          & 0.2477    & 0.2109    & 0.4194  \\
        Solar-pro2      & 0.0041    & 0.0485    & 0.1780  \\
        \addlinespace[2pt]
        LLaMA-3.1(8B)       & 0.0001    & 0.0923    & 0.1950  \\
        Mistral(7B)         & 0.0003    & 0.0044    & 0.1603  \\
        Qwen(8B)            & 0.0134    & 0.1136    & 0.0574  \\
        T5(2B)              & 0.0009    & 0.0037    & 0.1442  \\
      \bottomrule
    \end{tabular}
    \caption{Board Authorities}
    \label{tab:board-authorities_main_result}
  \end{subtable}

  % ===== 전체 캡션 =====
  \caption{Exact Match, Macro-F1 and Micro-F1 scores of Issue Type and Board Authorities classification}
  \label{tab:two-task_issue-board}
\end{table*}

\section{Tasks}

In this section, we formalize the benchmark’s three classification tasks in alignment with the IRAC framework. While we follow IRAC’s logical order, the tasks are defined as independent evaluation units without dependencies across them. A uniform input and leakage-prevention policy applies: to avoid answer leakage, we exclude all PTAB's opinion text, and by default inputs consist only of the \texttt{appellant\_arguments} and \texttt{examiner\_findings} produced by the Opinion Split.

We note that the benchmark does not include a task corresponding to the Application stage of IRAC. Application requires multi-step reasoning that connects legal rules to case-specific facts, which goes beyond the scope of classification. In this work, we focus on classification tasks as a first step, and leave Application to future research, where it can be more appropriately modeled through generation tasks that capture complex legal reasoning.

\subsection{Issue Type (IRAC–Issue)}
This task identifies which statutory grounds are disputed in a case. The model must contrast and synthesize the competing arguments of the appellant and the examiner to determine the contested legal issues, corresponding directly to the Issue stage of IRAC. The task is formulated as multi-label classification at the case-level. For evaluation, we report three complementary metrics: Exact Match as an overall case-level measure, Macro-F1 to capture performance under label imbalance, and Micro-F1 to reflect overall distributional performance. Additional evaluation metrics are reported in Appendix~\ref{tab:issue-result-overall}.

\subsection{Board Authorities (IRAC–Rule)}
This task predicts which procedural provisions under 37~C.F.R. are cited as authority for the Board’s decision, given the parties’ arguments and evidence. This corresponds to the Rule stage of IRAC. Like the Issue Type task, this task is modeled as a case-level multi-label classification and evaluated using the same metrics: Exact Match, Macro-F1, Micro-F1. Other evaluation metrics are provided in the Appendix \ref{tab:board-result-overall}.

\subsection{Subdecision (IRAC–Conclusion)}
This task predicts the Board’s final outcome for an appeal. The model must integrate conflicting claim-level arguments and facts from both sides and select a single conclusion for the case, corresponding to the Conclusion stage of IRAC. The task is framed as multi-class classification. For evaluation, we report Accuracy as the baseline overall measure, Macro-F1 to account for class imbalance, and Weighted-F1 to reflect performance across the empirical label distribution. Other evaluation metrics, such as micro-F1, are reported in the Appendix \ref{tab:subdecision-fine-result-overall} and \ref{tab:subdecision-coarse-result-overall}.

\begin{table*}[t]
  \centering
  \scriptsize
  % ===== 왼쪽 표 =====
  \begin{subtable}{0.48\textwidth}
    \centering
    \begin{tabular}{lccc}
      \toprule
      \textbf{Model} & \textbf{Accuracy} & \textbf{Macro-F1} & \textbf{Weighted-F1}\\
        \midrule
% ===================== BASE block =====================
        \multicolumn{4}{c}{\textbf{Split (Base)}} \\
        \cmidrule(lr){1-4}
        Claude-Sonnet-4 & 0.5658    & 0.1296    & 0.4854  \\
        Gemini-2.5-pro  & 0.5050    & 0.1635    & 0.4982  \\
        GPT-4o          & 0.4924    & 0.0997    & 0.4907  \\
        GPT-o3          & \textbf{0.5918}    & \textbf{0.1639}    & \textbf{0.5541}  \\
        Solar-pro2      & 0.5369    & 0.0779    & 0.3923  \\
        \addlinespace[2pt]
        LLaMA-3.1(8B)       & 0.4364    & 0.0767    & 0.4006  \\
        Mistral(7B)         & 0.1241    & 0.0251    & 0.1284  \\
        Qwen(8B)            & 0.4794    & 0.1024    & 0.4450  \\
        T5(2B)              & 0.0419    & 0.0142    & 0.0617  \\
        \midrule
% ===================== MERGE block =====================
        \multicolumn{4}{c}{\textbf{Merge}} \\
        \cmidrule(lr){1-4}
        Claude-Sonnet-4 & 0.5590    & 0.1129    & 0.4320  \\
        Gemini-2.5-pro  & 0.5114    & 0.1443    & 0.5036  \\
        GPT-4o          & 0.4592    & 0.0912    & 0.4353  \\
        GPT-o3          & \textbf{0.6086}    & \textbf{0.1683}    & \textbf{0.5682}  \\
        Solar-pro2      & 0.5420    & 0.0804    & 0.3932  \\
        \addlinespace[2pt]
        LLaMA-3.1(8B)       &0.5036     & 0.0696	& 0.0676  \\
        Mistral(7B)         &0.1265	    & 0.0572	& 0.0407  \\
        Qwen(8B)            & 0.4266    & 0.0698    & 0.4264    \\
        T5(2B)              & 0.0191	& 0.0794	& 0.0437  \\
        \midrule
% ===================== CLAIM block =====================
        \multicolumn{4}{c}{\textbf{Split+Claim}} \\
        \cmidrule(lr){1-4}
        Claude-Sonnet-4 & 0.5620    & 0.1272    & 0.4842  \\
        Gemini-2.5-pro  & 0.4908    & 0.4854    & 0.1433  \\
        GPT-4o          & 0.3804    & 0.0892    & 0.3581  \\
        GPT-o3          & \textbf{0.5884}	& \textbf{0.1692}	& \textbf{0.5538}  \\
        Solar-pro2      & 0.5373    & 0.0608    & 0.3966  \\
        \addlinespace[2pt]
        LLaMA-3.1(8B)       & 0.4125	& 0.0642	& 0.3938  \\
        Mistral(7B)         & 0.1209	& 0.0295	& 0.1205  \\
        Qwen(8B)            & 0.4368    & 0.0794    & 0.4364  \\
        T5(2B)              & 0.0225    & 0.0436    & 0.0168  \\
    \bottomrule
    \end{tabular}
    \caption{Subdecision (Fine-grained)}
    \label{tab:subdecision_fine_main_result}
  \end{subtable}\hfill
  % ===== 오른쪽 표 =====
  \begin{subtable}{0.48\textwidth}
    \centering
    \begin{tabular}{lccc}
      \toprule
      \textbf{Model} & \textbf{Accuracy} & \textbf{Macro-F1} & \textbf{Weighted-F1}\\
        \midrule
% ===================== BASE block =====================
        \multicolumn{4}{c}{\textbf{Split (Base)}} \\
        \cmidrule(lr){1-4}
        Claude-Sonnet-4 & 0.5625    & 0.2116    & 0.4900  \\
        Gemini-2.5-pro  & 0.5063    & \textbf{0.2366}    & 0.4927  \\
        GPT-4o          & 0.5045    & 0.2037    & 0.4863  \\
        GPT-o3          & \textbf{0.5863}    & 0.2126    & \textbf{0.5511}  \\
        Solar-pro2      & 0.5389    & 0.1356    & 0.3929  \\
        \addlinespace[2pt]
        LLaMA-3.1(8B)       & 0.4764    & 0.1551    & 0.4024  \\
        Mistral(7B)         & 0.0726    & 0.0758    & 0.0994  \\
        Qwen(8B)            & 0.4733    & 0.1692    & 0.4404  \\
        T5(2B)              & 0.0254    & 0.0499    & 0.0146  \\
        \midrule
% ===================== MERGE block =====================
        \multicolumn{4}{c}{\textbf{Merge}} \\
        \cmidrule(lr){1-4}
        Claude-Sonnet-4 & 0.5607	& 0.1788	& 0.4456  \\
        Gemini-2.5-pro  & 0.5119	& \textbf{0.2381}	& 0.5001  \\
        GPT-4o          & 0.4972	& 0.1820	& 0.4638  \\
        GPT-o3          & \textbf{0.6020}	& 0.2125	& \textbf{0.5631}  \\
        Solar-pro2      & 0.5423	& 0.1390    & 0.3967  \\
        \addlinespace[2pt]
        LLaMA-3.1(8B)       & 0.5229	& 0.1253	& 0.3922  \\
        Mistral(7B)         & 0.0823	& 0.0821    & 0.1168  \\
        Qwen(8B)            & 0.4163    & 0.1761    & 0.4223  \\
        T5(2B)              & 0.0234	& 0.0446	& 0.0092  \\
        \midrule
% ===================== CLAIM block =====================
        \multicolumn{4}{c}{\textbf{Split+Claim}} \\
        \cmidrule(lr){1-4}
        Claude-Sonnet-4 & 0.5639	& 0.2018	& 0.4889  \\
        Gemini-2.5-pro  & 0.4915    & 0.4840    & 0.2111  \\
        GPT-4o          &  0.3046	& 0.1206	& 0.2027  \\
        GPT-o3          &  \textbf{0.5783}	& \textbf{0.2068}    & \textbf{0.5426}  \\
        Solar-pro2      & 0.5364	& 0.1210	& 0.3977  \\
        \addlinespace[2pt]
        LLaMA-3.1(8B)       & 0.4741	& 0.1259	& 0.3909  \\
        Mistral(7B)         & 0.0587	& 0.0549    & 0.0721  \\
        Qwen(8B)            & 0.4605    & 0.1655    & 0.4439  \\
        T5(2B)              & 0.0136    & 0.0053    & 0.0142  \\
      \bottomrule
    \end{tabular}
    \caption{Subdecision (Coarse-grained)}
    \label{tab:subdecision_coarse_main_result}
  \end{subtable}

  % ===== 전체 캡션 =====
  \caption{Accuracy, Macro-F1 and Weighted-F1 scores of Subdecision (Fine-grained) and Subdecision (Coarse-grained) classification}
  \label{tab:two-tasks}
\end{table*}

\section{Experiments}

We describe the experimental setup, model lineup, and evaluation protocol for the three classification tasks. Unless otherwise noted, inputs are restricted to the \texttt{appellant\_arguments} and \texttt{examiner\_findings} obtained from the Opinion Split, with all PTAB’s opinion text excluded. For input-variation experiments, we compare three configurations under identical instructions: Split (Base), Merge, and Split+Claim. In the Split (Base) setting, appellant and examiner arguments are separated into distinct segments. Merge combines the two roles into a single role-neutral input, while Split+Claim augments the role-separated arguments with the patent’s claim text. These variants allow us to analyze the relative contributions of role signals (the distinction between appellant and examiner) and technical signals (the claim text) to model performance.

The model lineup includes five closed-source(commercial) LLMs and four open-source LLMs. The closed-source(commercial) models are Claude-Sonnet-4~\cite{claude}, Gemini-2.5-pro~\cite{Gemini}, GPT-4o, GPT-o3~\cite{GPTs}, and Solar-pro2~\cite{Solar}. The open-source models are LLaMA-3.1~\cite{Llama}, Mistral~\cite{Mistral}, Qwen~\cite{Qwen}, and T5~\cite{T5}. For closed-source(commercial) models, structured output features such as function calling were used to guarantee JSON-only responses. For open-source models, which lack native structured output capabilities, we enforced consistency by providing explicit format examples in the instruction and applying post-processing to convert outputs into valid JSON. This ensured that parsing errors were minimized across all runs.  

All tasks are evaluated in a zero-shot setting under a unified protocol. Detailed instruction templates, and prompts are provided in Appendix~\ref{sec:task-prompt} and model specifications are provided in the Appendix~\ref{sec:model} .

\section{Results}

We evaluate model performance across the three classification tasks, with task-level results reported in Tables~\ref{tab:issue-type_main_result}--\ref{tab:subdecision_coarse_main_result}; confusion heatmaps appear in the Appendix~\ref{sec:results}. Overall, closed-source(commercial) models consistently outperform open-source models, although all models exhibit limitations under long-tailed label distributions. Macro-F1 remains low across tasks, reflecting persistent difficulty with rare labels.  

\subsection{Closed-Source(commercial) vs. Open-Source Models}
As shown in the confusion heatmaps (Figures~\ref{fig:base_issue_type_all_heatmap}--\ref{fig:claim_subdecision_coarse_all_heatmap}), closed-source(commercial) models (Claude-Sonnet-4, Gemini-2.5-pro, GPT-4o, GPT-o3, Solar-pro2) achieve consistently higher accuracy and exhibit a stronger diagonal concentration, indicating greater reliability in classification performance. In the Issue Type task under the Split (Base) setting, closed-source(commercial) models reach Exact Match scores around 55--60\% with Micro-F1 scores close to 0.80, whereas open-source models are far less consistent: LLaMA-3.1 and Mistral remain below 35\% Exact Match, T5 collapses to below 10\%, and only Qwen approaches closed-source(commercial)-level performance. The Issue Type results thus provide the clearest illustration of the performance gap between closed-source(commercial) and open-source models.

\subsection{Input-Setting Effects}
Split (Base) provides the most reliable performance across tasks. Merge occasionally improves consistency for certain models, such as Claude-Sonnet-4 and GPT-o3, suggesting that role separation can sometimes introduce unnecessary variability. Split+Claim generally degrades performance: input length increases by roughly twice on average, and by a factor of three to four in terms of maximum token count, compared to Split (Base) (Table~\ref{tab:input_tokens}). This dilutes the salience of arguments and introduces irrelevant claim text as noise. The effect is most pronounced in the Board Authorities task (Table~\ref{tab:board-authorities_main_result}), where all models except Gemini-2.5-pro show a clear decline. Unlike Issue Type or Subdecision, which integrate technical facts with legal reasoning, Board Authorities is narrowly focused on mapping arguments to procedural rules. In this setting, claim text contributes little useful information and instead confuses the model, leading to a sharper performance drop. These results highlight that more input context is not uniformly beneficial: when tasks hinge primarily on legal rule alignment rather than technical content, excessive claim context may actively impair model reasoning.

\subsection{Invalid Response Patterns}
Another clear pattern, especially among open-source models, is the generation of labels outside the predefined set. For example, in Issue Type and Board Authorities tasks, models occasionally output arbitrary numbers or provisions not included in the label schema. This indicates both a failure to strictly follow instructions and a lack of domain alignment. Potential remedies include stronger prompt constraints (explicitly requiring outputs to be drawn only from the label set), post-filtering to reject out-of-label responses, and instruction tuning to reduce invalid or incomplete responses. Example cases of label deviations and invalid responses are presented in Appendix~\ref{sec:appendix-response-tendencies}.

\subsection{Summary}
Taken together, these results show that while closed-source(commercial) models can handle frequent labels and surface-level reasoning, all models struggle with long-tailed label distributions. The IRAC-based task design exposes these weaknesses across different stages, while the input-setting analysis underscores the importance of careful input design. Future work will build on these findings by exploring selective claim augmentation and instruction tuning as ways to improve alignment with PTAB-specific reasoning tasks.  

\section{Conclusion}

We presented PILOT-Bench, a benchmark to evaluate legal reasoning in the patent domain by aligning PTAB \textit{ex parte} appeals with USPTO patent data. By framing three IRAC-aligned classification tasks, we enable systematic assessment of LLMs’ ability to identify issues, map rules, and predict conclusions in appeal proceedings. Our experiments show that while closed-source(commercial) LLMs outperform open-source models, all models face persistent challenges with label imbalance and procedural-rule mapping. Input-variation analysis further demonstrates that simply adding all claims can harm performance, underscoring the need for more targeted data design.  

PILOT-Bench thus provides both a resource and an evaluation protocol to study how LLMs reason in a domain where technical detail and legal precision must be combined. We hope this benchmark will encourage further work at the intersection of NLP, law, and intellectual property.

\section{Future Work}

Beyond this study, we plan to pursue research-driven extensions of PILOT-Bench. A first direction is to expand beyond classification by introducing generation-based tasks that capture the IRAC Application stage, directly testing whether models can reason through the application of legal rules to facts. Second, we aim to explore selective claim augmentation and instruction tuning to mitigate noise and hallucination, thereby improving alignment with task constraints. Finally, we envision extending the benchmark to broader PTAB and USPTO contexts, enabling multi-procedure comparisons and richer evaluation of patent-domain legal reasoning.

\section*{Limitations}

This study has several limitations related to data collection and task design. First, the scope is restricted to PTAB \textit{ex parte} appeals, excluding AIA trial proceedings. While this aligns with source availability and our intended focus, it confines evaluation to appeal-centered cases. Second, although OCR quality is generally stable, no systematic, line-by-line correction against the source PDFs was performed; the converted text should not be regarded as a fully verified transcription. Similarly, the Opinion Split was generated solely via an LLM without human validation, so misclassifications may propagate into downstream tasks. Finally, the dataset exhibits substantial label imbalance. To address this, Subdecision outcomes were consolidated into six coarse labels via LLM-based normalization without additional rebalancing. Partnering with domain experts to vet and refine this schema may yield further gains in robustness and interpretability.

\section*{Ethical Considerations}
This benchmark is released for research purposes only and must not be used to automate, replace, or appear to provide legal advice or adjudicative decisions. All documents originate from public USPTO/PTAB sources; we redistribute only derived annotations/splits/metadata and remove any incidental PII found during OCR. Users remain responsible for compliance with applicable laws and professional standards. Model outputs may contain errors and require qualified human review.

\section*{Acknowledgments}
This research was supported by the MSIT (Ministry of Science and ICT), Korea, under grants through the National Research Foundation of Korea (NRF) (No. RS-2024-00354675, 70\%) and the ICAN (ICT Challenge and Advanced Network of HRD) support program supervised by the IITP (Institute for Information \& Communications Technology Planning \& Evaluation) (IITP-2023-RS-2023-00259806, 30\%)

\bibliography{custom}

\clearpage

\appendix
\section*{Appendix}
\label{sec:appendix}

\section{Data Card}
\begin{itemize}
\item \textbf{Licensing Information} The dataset is released under the Creative Commons Attribution 4.0 International License.
\item \textbf{Data Domain} Patent Domain
\item \textbf{Languages} The dataset contains English text only.
\item \textbf{Dataset Composition} PTAB OCR, PTAB Opinion Split, PTAB Metadata, and USPTO Structured Data.
\item \textbf{Computational Resources} Experiments were run on two RTX 4090(24GB) and two H100(80GB) GPUs
\end{itemize}

\section{Data Format and Structure}

\subsection{PTAB Decision}
Each PTAB decision is distributed as a JSON file named after the official decision filename (e.g., \texttt{2018004769\_DECISION.json}). We release two corpus variants: PTAB OCR and PTAB Opinion Split. PTAB OCR provides page-level Optical Character Recognition (OCR) text, providing extracted from each decision. PTAB Opinion Split segments the decision text into four categories: \texttt{appellant\_arguments}, \texttt{examiner\_findings}, \texttt{ptab\_opinion}, and \texttt{facts}.

\subsection{PTAB Metadata}
we release a PTAB Metadata JSON aligned PTAB decision JSON files. PTAB Metadata contains 35 fields per decision, including the targets used in our classification tasks: \texttt{issueType}, \texttt{boardRulings}, and \texttt{subdecisionTypeCategory}. Table~\ref{tab:ptab-metadata} shows the metadata JSON fields.

\subsection{USPTO Structured Data}
For each decision, we include the corresponding USPTO patent data as a single JSON file within the directory for that PTAB Decision filename, named by the patent’s application or publication number (e.g., \texttt{2018004769\_DECISION/US20140127537A1.json}).

\section{Dataset Creation}
\subsection{Source Data}
We collected 25,829 PTAB decisions (1993–2024) and 176,627 metadata records (1997–2025) via the PTAB API v2\footnote{\url{https://developer.uspto.gov/api-catalog/ptab-api-v2}}. We also retrieved patent full texts and bibliographic metadata from USPTO Bulk Data\footnote{\url{https://data.uspto.gov/bulkdata/datasets}}, covering 2006–2024.

\subsection{Patent-Term Filtering}
Considering the statutory patent term (typically 20 years from the filing date), we restrict our analysis to PTAB decisions dated 2006 or later, yielding 22,439 cases.

\subsection{OCR Quality Filtering}
We require page-level OCR for decision text analysis. Nonstandard layouts—often due to missing cover pages—disrupted caption normalization and section detection. To stabilize OCR, we retain only decisions with a cover page, resulting in 18,738 cases.

\subsection{Case-Thread Normalization}
We define the analysis scope for \textit{ex parte} appeal case threads and apply metadata-driven preprocessing to normalize threads and remove duplicates. To ensure a reproducible one-to-one mapping between each case and its associated patent text, we adopt a single target per case and restrict the analysis to a subset of procedural variants. Records that could yield duplicate or ambiguous labels are excluded.

\begin{itemize}
    \item \textbf{Exact duplicates} Decision records  Decision records that are identical across all fields; a single canonical decision record is retained.
    \item \textbf{Application number / document name duplicates} 
    When multiple decision records share \texttt{documentName} and \texttt{appellantApplicationNumberText}, we reconcile the PTAB Decision with PTAB Metadata and preserve one consistent decision record.
    \item \textbf{Subsequent proceedings (rehearing/reconsideration/reexamination)} 
    Subsequent decisions within the same proceeding can produce multiple decision records for a single dispute. we retain one representative decision record per (\texttt{documentName}, \texttt{decisionDate}) pair.
    \item \textbf{Separate opinions (dissent/concurring)} Separately authored opinions are excluded because they may introduce competing rationales and thus ambiguous case-level labels. Only the unified decision record is kept for downstream tasks.
\end{itemize}

\subsection{OCR Parsing}
From the OCR text, we removed cover-page bibliographic fields (e.g., \texttt{Application No.}, \texttt{Filing Date}, \texttt{First Named Inventor}) that duplicate metadata entries, thereby preventing redundancy. To maintain linguistic consistency and improve OCR robustness, we also removed non-English text.

\subsection{Section Segmentation}
To support a logical decomposition of each decision, we defined a header dictionary comprising \texttt{DECISION ON APPEAL}, \texttt{STATEMENT OF THE CASE}, \texttt{ANALYSIS}, \texttt{DECISION/ORDER}, and \texttt{FOOTNOTES}, and we then performed section-level segmentation using GPT-o3 (3-2025-04-16). Decisions in which \texttt{STATEMENT OF THE CASE} or \texttt{ANALYSIS} could not be extracted—e.g., dismissals following a Request for Continued Examination (RCE) or express abandonment—were excluded from the analysis.

\subsection{PTAB Opinion Split}
Using the primary reasoning sections \texttt{STATEMENT OF THE CASE} and \texttt{ANALYSIS} as input, we split each decision with gemini-2.5-pro into four categories: \texttt{appellant\_arguments}, \texttt{examiner\_findings}, \texttt{ptab\_opinion}, and \texttt{facts}. Only \texttt{appellant\_arguments} and \texttt{examiner\_findings} are used as inputs to downstream tasks. Figure \ref{fig:opinion_split_prompt} presents the prompt for opinion splitting.

\subsection{PTAB to USPTO Mapping}
We align PTAB decision records with USPTO patent records via the application number, matching PTAB \texttt{appellantApplicationNumberText} to USPTO \texttt{application-reference/doc-number}. When a single application number is associated with multiple publications, we select one representative publication anchored to the PTAB \texttt{decisionDate}. Applications predating 2006 fall outside the coverage of our USPTO corpus and are omitted. This alignment yields 15,482 PTAB--USPTO links.

\subsection{USPTO Structured Data}
To preserve claim dependencies, each claim carries a \texttt{depend\_on} pointer to its parent claim. We further factor claim text into component-level units and arrange them hierarchically to support granular analyses in subsequent work. Figure~\ref{fig:uspto} depicts the schema.

\section{Classification Tasks}

\subsection{Prediction Targets}
Our tasks comprise three targets: issue type, board authorities, and subdecision. For consistency in evaluation, instances with missing \texttt{Board Authorities} (empty) are systematically mapped to \texttt{Others} label. 

\subsection{Label Details}
\label{sec:label}
Table \ref{tab:appendix-issue-def}--\ref{tab:subdecision-coarse-mapping-others} enumerates the full labels used in our experiments and their definitions.

\subsection{Prompt}
\label{sec:task-prompt}
Figure \ref{fig:issue_prompt}--\ref{fig:subdecision_prompt} are the prompts used for each task; Issue Type, Board Authorities, Subdeicision (Fine/Coarse).

\section{Statistics and Analysis}
\subsection{Input Tokens per Variants}
Table~\ref{tab:input_tokens} reports the average and maximum input token counts per input variant for the Board Authorities task, measured with the Gemini tokenizer.

\subsection{Experiment Results}
\label{sec:results}
Tables~\ref{tab:issue-result-overall}–\ref{tab:subdecision-coarse-result-overall} present results for all evaluation metrics. Table~\ref{tab:issue-result-overall} shows that T5 attains unusually high recall despite weaker Exact Match, Micro-F1, and Macro-F1. Inspection of Figure~\ref{fig:t5-base}-\ref{fig:t5-claim} reveals a systematic tendency to emit the full five-label set (\texttt{[101,102,103,112,Others]}), which mechanically inflates recall in the multi-label setting by covering most labels while simultaneously depressing precision and exact match. All models’ confusion heatmaps can be found in Figures~\ref{fig:base_issue_type_all_heatmap}--\ref{fig:claim_subdecision_coarse_all_heatmap}

\subsection{PTAB Subproceeding Types by Year}
\label{sec:appendix-subproceeding}

To illustrate the oral distribution and procedural composition of the PTAB corpus, 
we analyzed the number of decisions per year and subproceeding type (\textit{REEXAM}, \textit{REGULAR}, and \textit{REISSUE}) 
based on the PTAB Document JSON metadata. 
Figure~\ref{fig:subproceeding-per-year} and Table~\ref{tab:subproceeding-type} 
show a steady increase in \textit{REGULAR} appeal decisions from 2010 to 2017, 
followed by a gradual decline consistent with overall PTAB appeal volume trends. 
\textit{REEXAM} and \textit{REISSUE} proceedings account for less than 5\% of total decisions, 
confirming that the dataset is dominated by regular \textit{ex parte} appeals—the intended focus of PILOT-Bench.

\subsection{Document Length Statistics of Opinion Split Data}
\label{sec:appendix-length-stats}
We provide document and role aspect descriptive statistics to quantify the scale and variability of the Opinion Split data. 
Table~\ref{tab:split-length} summarizes the word-level statistics, and Table~\ref{tab:split-length-char} presents the corresponding character-level statistics. 
These results show that PTAB \textit{ex parte} decisions vary widely in length, with the \textit{Analysis} section dominating the total word count and the split inputs maintaining a balanced representation of opposing arguments.

\subsection{Linked Patents per PTAB Case}
\label{sec:appendix-linked-case}

To quantify the connectivity between PTAB decisions and their associated patents, 
we examined the number of linked patents per case after PTAB–USPTO alignment. 
Each PTAB case contains one \textit{base patent} (the appellant’s patent) 
and zero or more \textit{prior patents} cited as prior art or reference patents in the appeal record. 
Figure~\ref{fig:linked_case_dist} and Figure~\ref{fig:linked_case_year} 
visualize the distribution of linked patents across cases and its yearly trend.

On average, each PTAB case is connected to approximately \textbf{2.05 patents}, 
consisting of one base patent and roughly one additional prior patent. 
The average base-to-prior ratio is about \textbf{0.64}, 
indicating that while most cases are linked to a single prior reference, 
a small number of cases involve more complex prior-art networks (up to 14 linked patents). 
Table~\ref{tab:linked_case_stats} reports detailed summary statistics.

\section{Model}
\label{sec:model}
This study evaluates both closed-source(commercial) and open-source models. For the open-source group, we primarily used small models in the 2B–8B parameter range due to computational constraints. We expect larger variants of the same architectures (>8B parameters) and models with dedicated reasoning modes to achieve higher performance. Details on model sizes are provided below.

\begin{itemize}
    \item \textbf{Closed-source(commercial) Models} gpt-4o-2024-08-06, gpt-o3-2025-04-16, claude-sonnet-4-20250514, gemini-2.5-pro, solar-pro2-250710
    \item \textbf{Open-source Models} Llama-3.1-8B-Instruct, Qwen3-8B, Mistral-7B-Instruct-v0.3, t5gemma-2b-2b-ul2-it
\end{itemize}

\subsection{Post-Processing of Model Outputs}
For open-source models, we instructed JSON only output at the prompt stage. In practice, some responses exhibited formatting errors, so we applied content-preserving normalization. Specifically, (i) we corrected parsing errors caused by missing or superfluous brackets or quotation marks with minimal edits, (ii) we restored character-level fragmented outputs (e.g., “{”, “i”, “s”, “s”, “u”, …) to valid contiguous strings, and (iii) we removed duplicated labels such as “103”, “103”, “103”. This pipeline was designed to enforce schema consistency without altering the meaning of the original responses.

\subsection{Response Tendencies}
\label{sec:appendix-response-tendencies}
\subsubsection{Closed-Source(commercial) Models}
\begin{itemize}
    \item \textbf{Issue Type} Claude intermittently returned \texttt{<UNKNOWN>}.
    \item \textbf{Board Authorities} According to the labels, citations such as \texttt{37 CFR 1.104}, \texttt{37 CFR 1.111}, \texttt{37 CFR 41.37(c)(iv)} should be assigned to \texttt{Others}; nevertheless, the model occasionally emitted them as distinct labels.
\end{itemize}

\subsubsection{Open-Source Models}
\begin{itemize}
    \item \textbf{Issue Type} We observed frequent deviations from the label set, bare numerals (e.g., \texttt{51}, \texttt{22}); subsection-annotated variants (e.g., \texttt{102(b)}, \texttt{103(a)}, \texttt{102(e)} instead of base labels \texttt{102}, \texttt{103}); and unstructured natural language text (e.g., “The Examiner found that claims …”).
    \item \textbf{Board Authorities} Category confusions and hallucinated citations were common. Statutory grounds intended for the Issue Type task (e.g., \texttt{35 U.S.C. § 103(a)}, \texttt{35 U.S.C. § 102(b)}) were misassigned as Board Authorities. Provisions outside our label set (e.g., \texttt{37~C.F.R. §~41.37(c)(1)(ii)})—which should map to \texttt{Others}—were emitted as labels. We also observed nonexistent citations in our dataset (e.g., \texttt{37~C.F.R. §~41.132}, \texttt{§~101}, \texttt{§~102(e)}).
    \item \textbf{Subdecision} Mistral tended to produce natural language text rather than  schema labels (e.g., “Claims 1–3, 17–23, 25, and 28–30 stand rejected.”).
\end{itemize}

\subsection{Evaluation Protocol and Response Rates}
\subsubsection{Evaluation Protocol} By default, we evaluated 15,482 cases. For each model–task pair, we allowed up to ten retries. A case was marked as a \texttt{non-answer} if (i) no output was produced, (ii) the model provided a rationale without a final label, or (iii) the input text was echoed verbatim or the response consisted of repetitive content.

\subsubsection{Response Rates}
\begin{itemize}
    \item \textbf{Solar-pro2} Owing to maximum context-length limits, evaluation under Split+Claim covered 15,481 samples. See Table~\ref{tab:input_tokens} for average input length.
    \item \textbf{T5} Under the Base and Merged, evaluations of Subdecision-Fine and Subdecision-Coarse yielded on average 15,470 valid responses. Despite up to ten retries, we frequently observed outputs consisting only of explanatory text without a label or terminating in repetitive content. Under Split+Claim, response rates declined across all tasks, with non-answers increasing via partial claim echoes or verbatim reproductions of the input; accordingly, metrics for Split+Claim were computed on approximately 15,040 samples.
    \item \textbf{Mistral.} Under Split+Claim for Board Authorities, the model frequently returned the input verbatim. Evaluation proceeded with 15,481 samples.
    
\end{itemize}

% @@@@@@@@@@@@@@@@@@@@@@@@@@@@@@@@@@@@@@@@@@@
% @@@@@@@@@@ Figure, Table Zone @@@@@@@@@@@@@
% @@@@@@@@@@@@@@@@@@@@@@@@@@@@@@@@@@@@@@@@@@@

\begin{table*}[t]
  \centering
  \small
  \setlength{\tabcolsep}{5pt}
    \begin{tabularx}{\textwidth}{@{} S D E @{}}
    \toprule
    \textbf{Name} & \textbf{Definition} & \textbf{Example} \\
    \midrule
        proceedingNumber & PTAB proceeding ID & \texttt{2018004769} \\
        decisionTypeCategory & Decision type & \texttt{"Decision"} \\
        subdecisionTypeCategory & Final outcome of decision & \texttt{“Affirmed”} \\
        documentName & Decision PDF filename & \texttt{“2018004769\_DECISION.pdf”} \\
        proceedingTypeCategory & Proceeding type & \texttt{“Appeal”} \\
        subproceedingTypeCategory & Sub-type of proceeding & \texttt{“REGULAR”} \\
        documentIdentifier & Document ID & \texttt{“201800476914127348Appeal ...“} \\
        objectUuId & Internal repository ID & \texttt{“workspace: ...“} \\
        respondentTechnologyCenterNumber & Respondent USPTO Technology Center(TC) & \texttt{“1700”} \\
        respondentPartyName & Respondent party name & \texttt{“Samsung SDI Co., Ltd. et al”} \\
        respondentGroupArtUnitNumber & Respondent Group Art Unit(GAU) number & \texttt{“1727”} \\
        respondentPatentNumber & Respondent patent number & \texttt{“10028104”} \\
        respondentApplicationNumberText & Respondent application number & \texttt{14127348} \\
        appellantTechnologyCenterNumber & Appellant USPTO Technology Center(TC) & \texttt{“1700"} \\
        appellantPatentOwnerName & Appellant name & \texttt{“Samsung SDI Co., Ltd. et al”} \\
        appellantPartyName & Appellant party name & \texttt{“Samsung SDI Co., Ltd. et al”} \\
        appellantGroupArtUnitNumber & Appellant Group Art Unit(GAU) number & \texttt{“1727”} \\
        appellantInventorName & Appellant inventor(s) name & \texttt{“Claus Gerald Pflueger et al”} \\
        appellantCounselName & Appellant Counsel/firm & \texttt{“Maginot, Moore \& Beck\ \,LLP”} \\
        appellantGrantDate & Appellant patent grant date & \texttt{“03-27-2018”} \\
        appellantPatentNumber & Appellant patent number & \texttt{“9925542”} \\
        appellantApplicationNumberText & Appellant application number. & \texttt{14127348} \\
        appellantPublicationDate & Appellant publication date & \texttt{“05-08-2014”} \\
        appellantPublicationNumber & Appellant publication number & \texttt{“20140127537A1"} \\
        ocrSearchText & OCR text by USPTO & \texttt{“14127348,Patent\_Board ...“} \\
        issueType & Statutory sections under 35~U.S.C. & \texttt{[“103”]} \\
        boardRulings & Regulatory provisions cited & \texttt{[“35 USC 134"]} \\
        decisionDate & Decision date & \texttt{“03-21-2019”} \\
        documentFilingDate & Filing date of the decision doc & \texttt{“03-21-2019"} \\
        thirdPartyName & Third party name & \texttt{“SMITH \&  NEPHEW, INC.”} \\
        file\_name & Basename without extension. & \texttt{“2018004769\_DECISION”} \\
        \textbf{issueType\_label} & Label of Issue Type task & \texttt{[“103”]} \\
        \textbf{boardAuthorities\_label} & Label of Board Authorities task & \texttt{[Others]} \\
        \textbf{subdecisionType\_label} & Fine-grained label of Subdeicision task & \texttt{“Affirmed”} \\
        \textbf{subdecisionTypeCoarse\_label} & Coarse-grained label of Subdeicision task & \texttt{“Affirmed”} \\
        \bottomrule
    \end{tabularx}
  \caption{PTAB metadata fields}
  \label{tab:ptab-metadata}
\end{table*}

% -------

\begin{figure*}[t]
    \centering
    \includegraphics[width=0.9\linewidth]{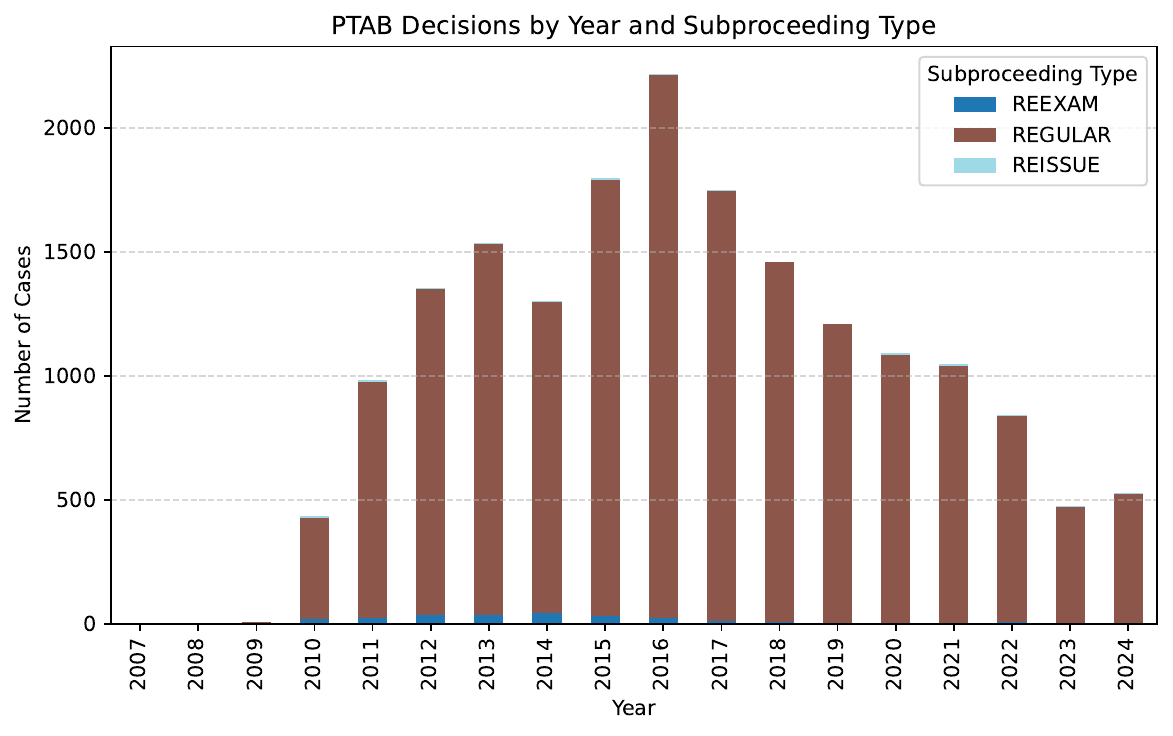}
    \caption{PTAB decisions by year and subproceeding type (2007–2024).}
    \label{fig:subproceeding-per-year}
\end{figure*}

% -----

\begin{table*}[t]
\centering
\small
\setlength{\tabcolsep}{6pt}
\begin{tabular}{lccc}
\toprule
\textbf{Year} & \textbf{REEXAM} & \textbf{REGULAR} & \textbf{REISSUE} \\
\midrule
2007 & 1  & 0   & 0 \\
2008 & 0  & 1   & 0 \\
2009 & 0  & 9   & 0 \\
2010 & 19 & 410 & 7 \\
2011 & 25 & 949 & 11 \\
2012 & 36 & 1314 & 6 \\
2013 & 35 & 1498 & 4 \\
2014 & 44 & 1256 & 4 \\
2015 & 34 & 1758 & 5 \\
2016 & 25 & 2192 & 1 \\
2017 & 14 & 1734 & 2 \\
2018 & 8  & 1452 & 0 \\
2019 & 5  & 1205 & 0 \\
2020 & 6  & 1078 & 7 \\
2021 & 4  & 1038 & 6 \\
2022 & 7  & 830  & 6 \\
2023 & 5  & 469  & 1 \\
2024 & 6  & 518  & 3 \\
\bottomrule
\end{tabular}
\caption{Number of PTAB decisions by subproceeding type from 2007 to 2024.}
\label{tab:subproceeding-type}
\end{table*}

% ---------

\begin{table*}[t]
\centering
\small
\setlength{\tabcolsep}{6pt}
\begin{tabular}{lcccccc}
\toprule
\textbf{Section / Role} & \textbf{Count} & \textbf{Mean (Words)} & \textbf{Median} & \textbf{Std} & \textbf{Min} & \textbf{Max} \\
\midrule
\textbf{Overall (Pre-Split)}   & 18,049 & 1,864.3 & 1,551 & 1,143.6 & 0 & 10,261 \\
\textit{Statement of the Case} & 17,919 & 433.4 & 366 & 276.5 & 19 & 4,685 \\
\textit{Analysis}              & 18,042 & 1,434.5 & 1,130 & 1,064.9 & 9 & 9,764 \\
\midrule
\textbf{Overall (Post-Split)}  & 18,049 & 1,409.1 & 1,173 & 935.7 & 0 & 10,039 \\
\textsf{appellant\_arguments}  & 17,445 & 296.5 & 235 & 242.6 & 3 & 2,613 \\
\textsf{examiner\_findings}    & 17,766 & 306.7 & 248 & 239.4 & 10 & 2,827 \\
\textsf{ptab\_opinion}         & 18,041 & 821.0 & 634 & 674.2 & 5 & 8,532 \\
\bottomrule
\end{tabular}
\caption{Descriptive statistics of document and role-level word counts in the PTAB Opinion Split dataset.}
\label{tab:split-length}
\end{table*}

\begin{table*}[t]
\centering
\small
\setlength{\tabcolsep}{6pt}
\begin{tabular}{lcccccc}
\toprule
\textbf{Section / Role} & \textbf{Count} & \textbf{Mean (Chars)} & \textbf{Median} & \textbf{Std} & \textbf{Min} & \textbf{Max} \\
\midrule
\textbf{Overall (Pre-Split)}   & 18,049 & 11,565.6 & 9,563 & 7,202.5 & 1 & 64,872 \\
\textit{Statement of the Case} & 17,919 & 2,690.3 & 2,241 & 1,749.8 & 120 & 28,950 \\
\textit{Analysis}              & 18,042 & 8,875.3 & 7,126 & 6,730.4 & 85 & 62,180 \\
\midrule
\textbf{Overall (Post-Split)}  & 18,049 & 8,748.5 & 7,245 & 5,883.9 & 2 & 64,594 \\
\textsf{appellant\_arguments}  & 17,445 & 1,856.2 & 1,468 & 1,525.4 & 14 & 17,163 \\
\textsf{examiner\_findings}    & 17,766 & 1,876.9 & 1,511 & 1,475.3 & 53 & 17,486 \\
\textsf{ptab\_opinion}         & 18,041 & 5,107.2 & 3,926 & 4,250.6 & 30 & 54,854 \\
\bottomrule
\end{tabular}
\caption{Descriptive statistics of document and role-level character counts in the PTAB Opinion Split dataset.}
\label{tab:split-length-char}
\end{table*}
% -----

\begin{figure*}[t]
    \centering
    \includegraphics[width=0.5\linewidth]{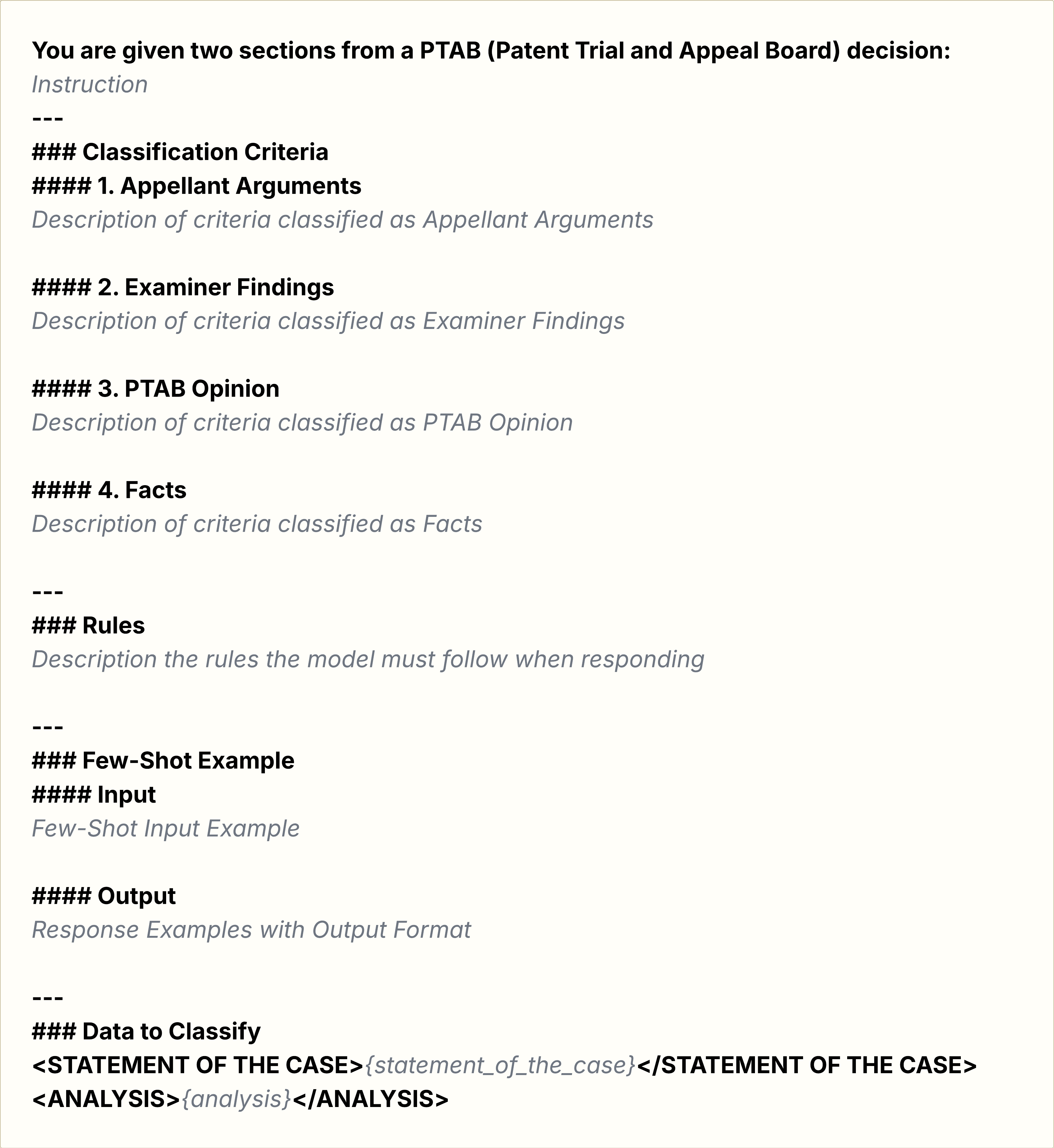}
    \caption{Opinion Split prompt construction}
    \label{fig:opinion_split_prompt}
\end{figure*}

\begin{figure*}[t]
  \includegraphics[width=\linewidth]{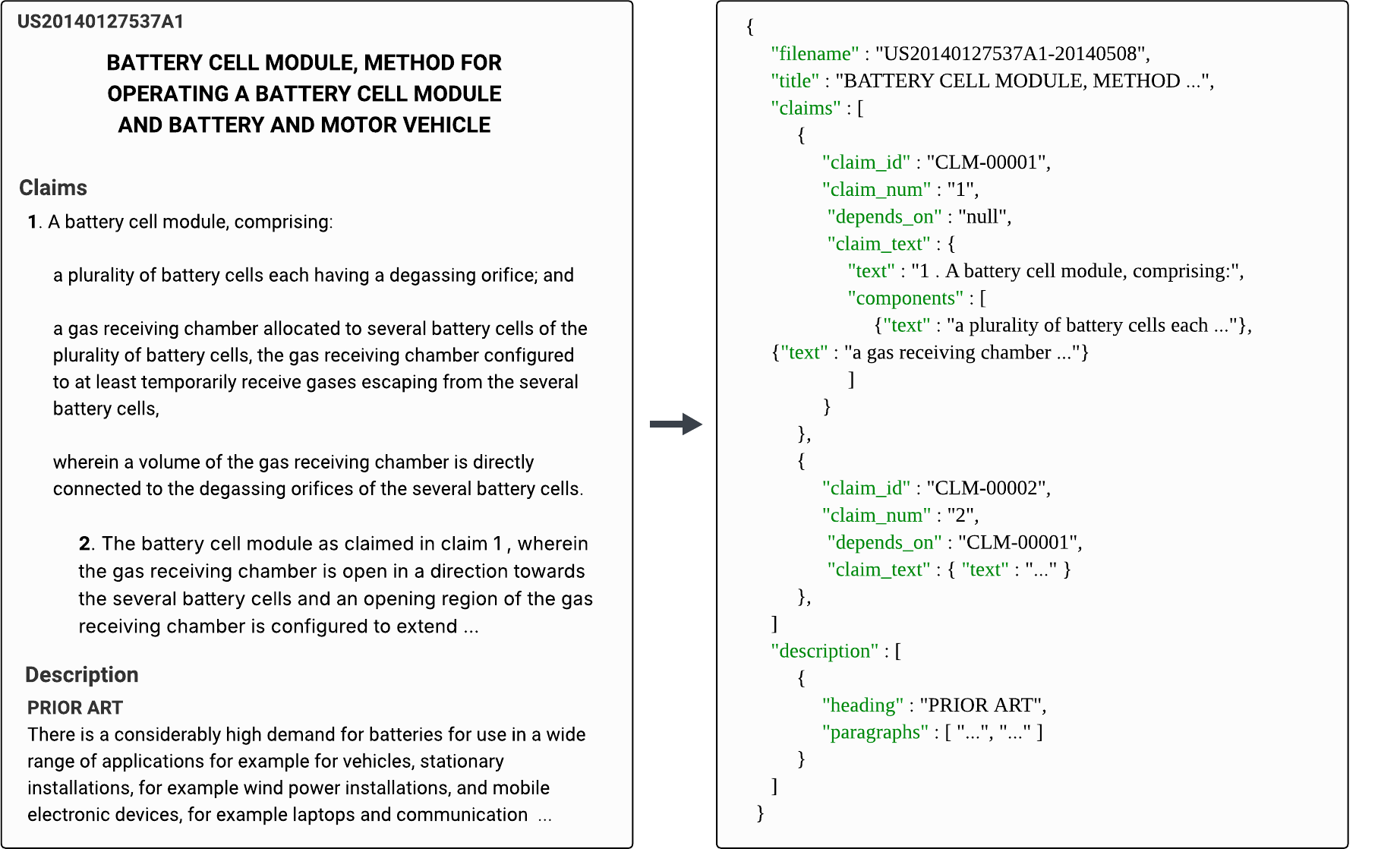}
  \caption{USPTO Structured Data structure}
  \label{fig:uspto}
\end{figure*}

\begin{figure*}[t]
    \centering
    \includegraphics[width=0.5\linewidth]{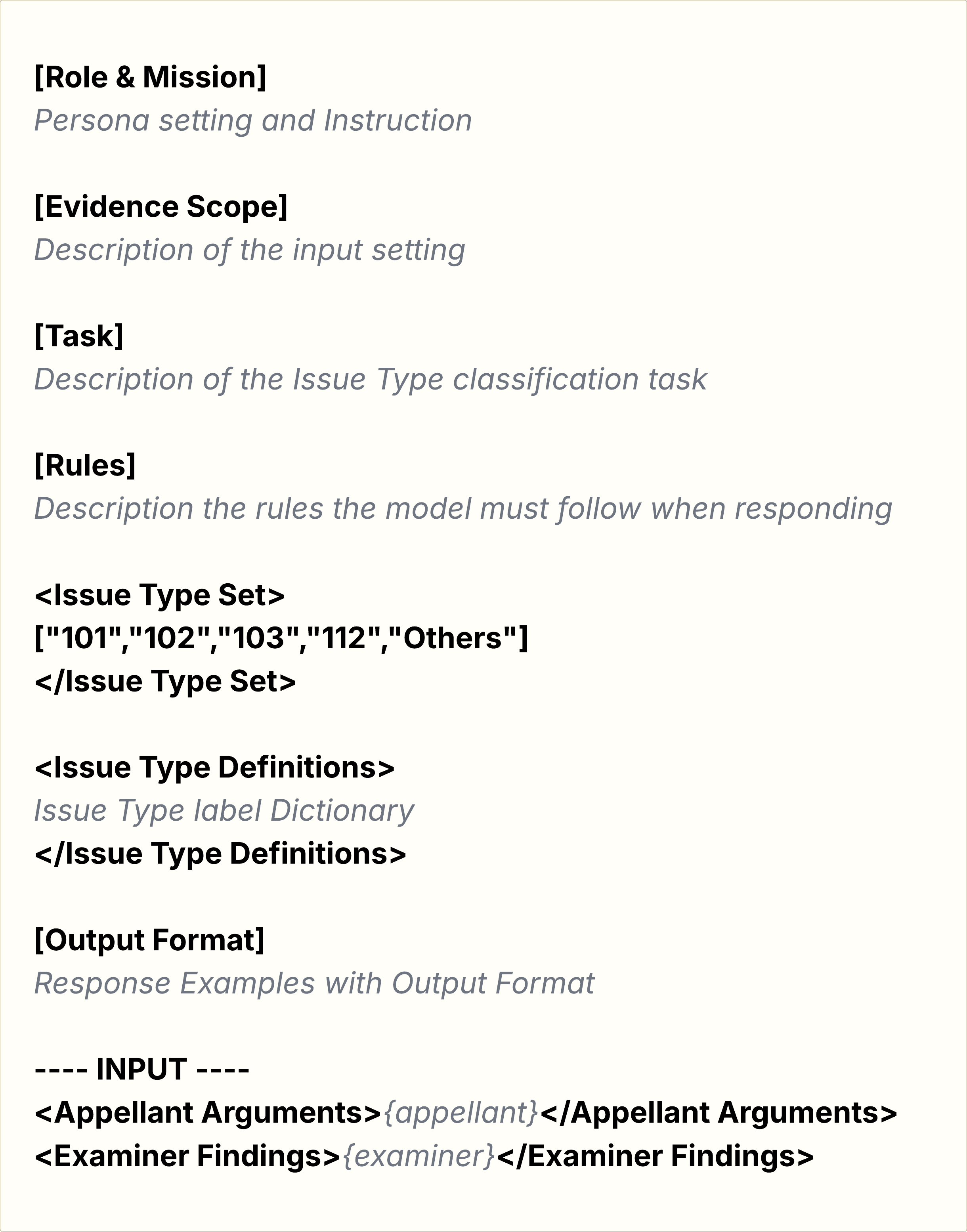}
    \caption{Issue Type classification prompt construction}
    \label{fig:issue_prompt}
\end{figure*}

\begin{figure*}[t]
    \centering
    \includegraphics[width=0.5\linewidth]{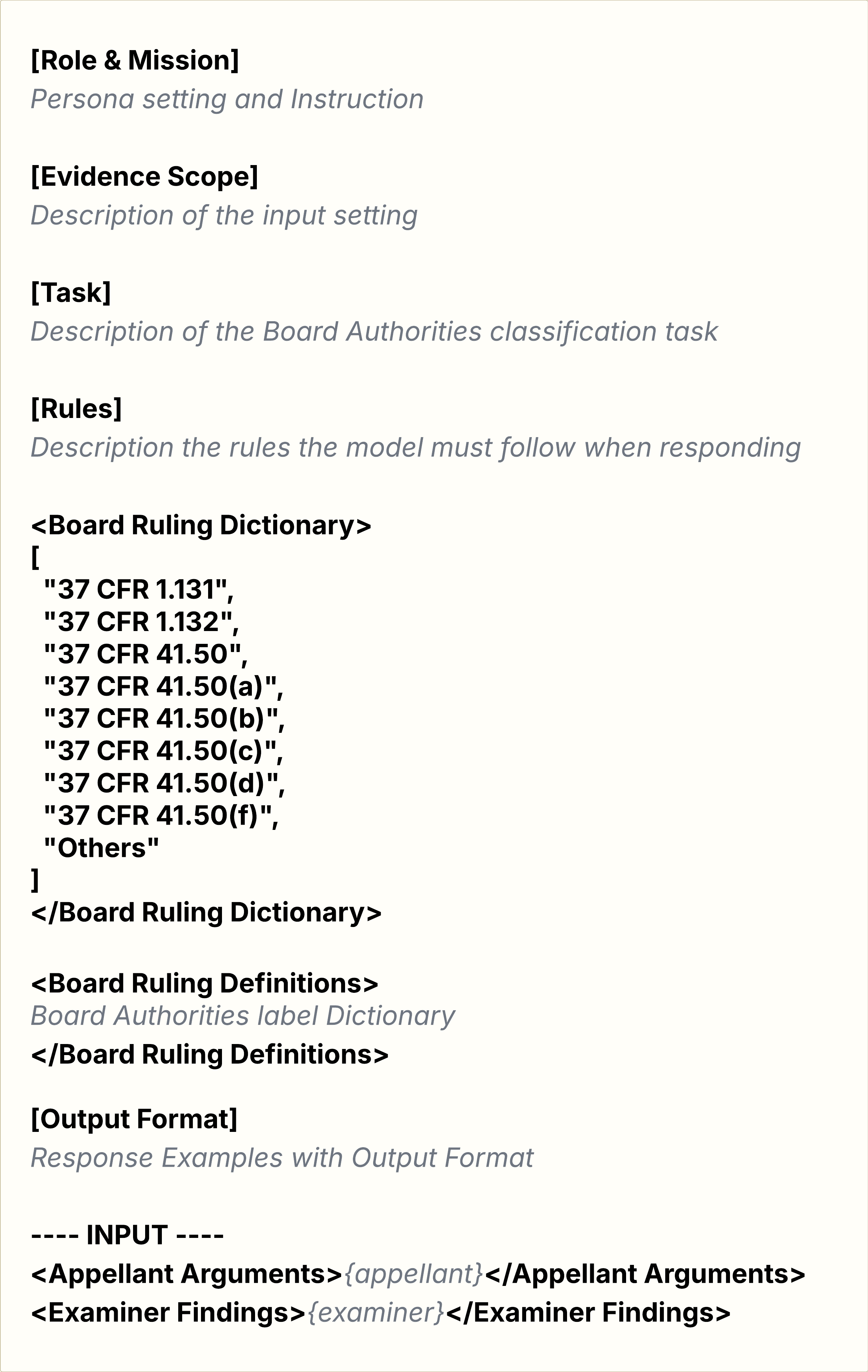}
    \caption{Board Authorities classification prompt construction}
    \label{fig:board_prompt}
\end{figure*}

\begin{figure*}[t]
    \centering
    \includegraphics[width=0.5\linewidth]{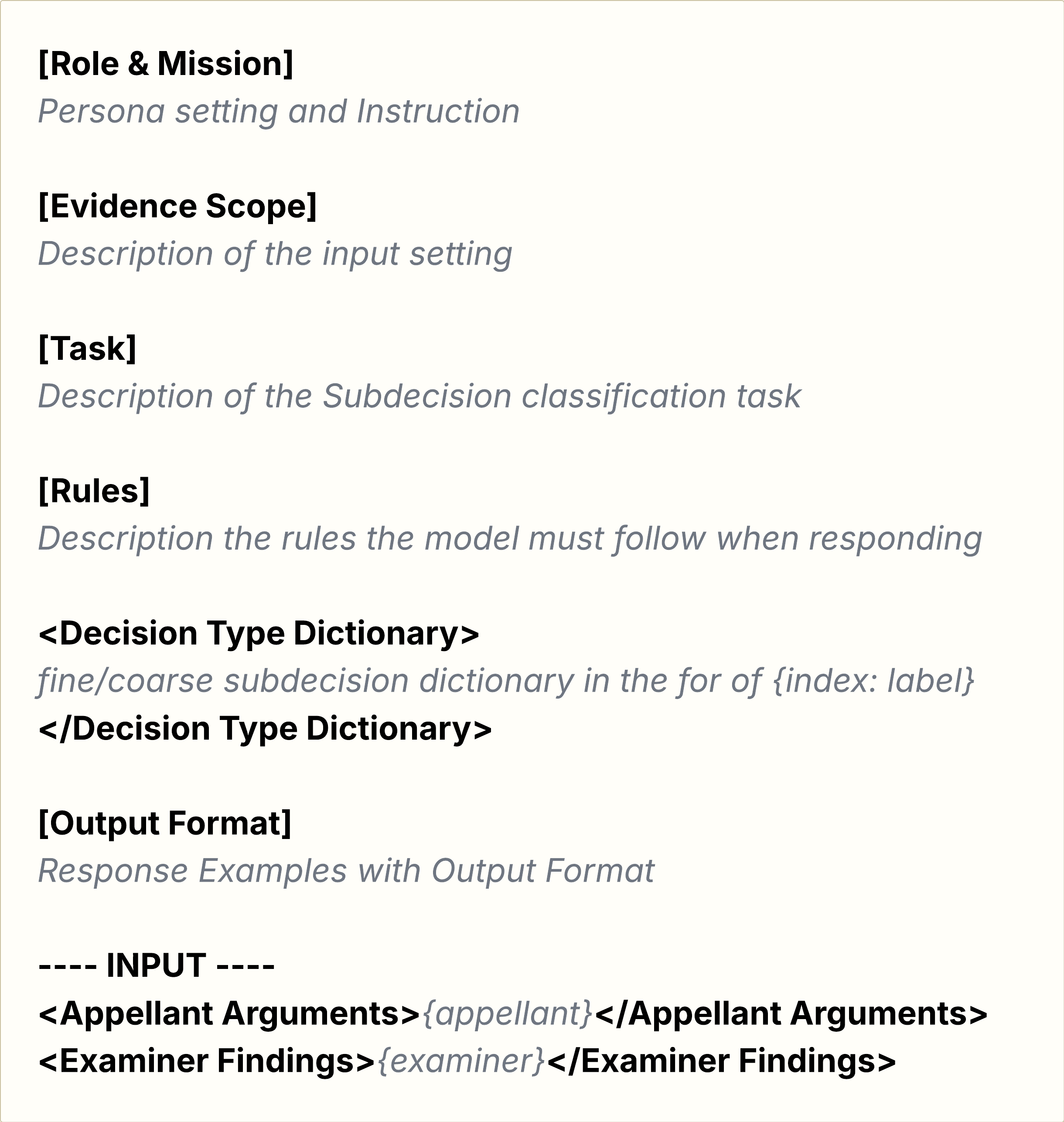}
    \caption{Subdecision (Fine/Coarse) classification prompt construction}
    \label{fig:subdecision_prompt}
\end{figure*}

\begin{table*}[t]
    \centering
    \small
    \setlength{\tabcolsep}{4pt} % 칸 간격(선택)
    \begin{tabular}{@{}p{0.24\linewidth} p{0.24\linewidth} p{0.24\linewidth} p{0.24\linewidth}@{}}
    \hline
    \textbf{Statistic} & \textbf{Split (Base)} & \textbf{Merge} & \textbf{Split+Claim} \\
    \hline
    Average & 2026.14 & 1730.00 & 4876.58 \\
    Maximum & 6109.00 & 5193.00 & 20924.00 \\
    \hline
\end{tabular}
\caption{Average and Maximum input tokens by variant (Board Authorities; Gemini tokenizer)}
\label{tab:input_tokens}
\end{table*}

% -------
\begin{table*}[t]
\centering
\small
\setlength{\tabcolsep}{5pt}
\begin{tabular}{lccc}
\toprule
\textbf{Statistic} & \textbf{Base Count} & \textbf{Prior Count} & \textbf{Total} \\
\midrule
Count & 78{,}480 & 78{,}480 & 78{,}480 \\
Mean & 0.99 & 1.06 & 2.05 \\
Std. Dev. & 0.10 & 1.47 & 1.47 \\
Min & 0 & 0 & 1 \\
Max & 1 & 13 & 14 \\
\bottomrule
\end{tabular}
\caption{Summary statistics of linked patents per PTAB case. Each case contains one base patent and zero or more prior patents.}
\label{tab:linked_case_stats}
\end{table*}
% ------------

% ------------

\begin{figure*}[t]
    \centering
    \includegraphics[width=0.9\linewidth]{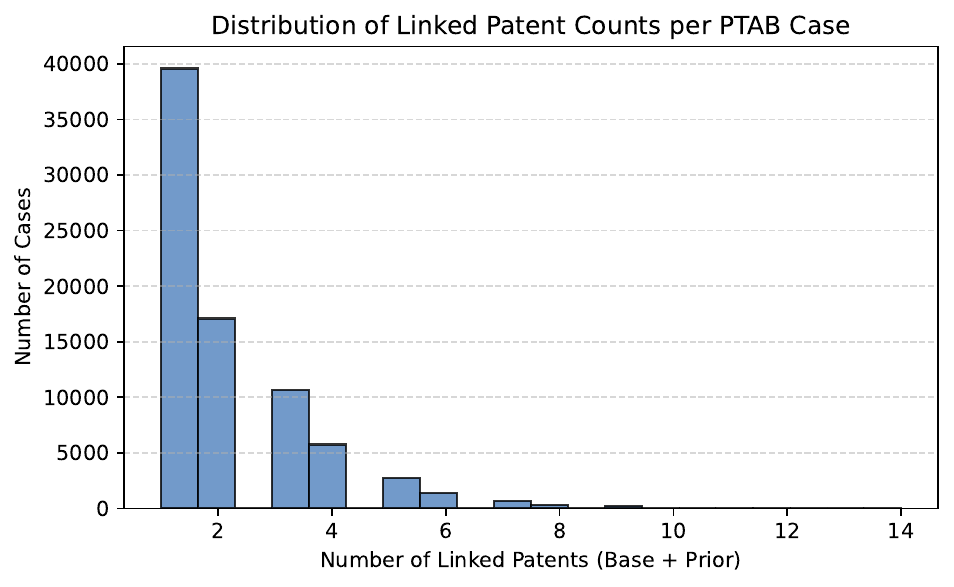}
    \caption{Distribution of the number of linked patents (base + prior) per PTAB case.}
    \label{fig:linked_case_dist}
\end{figure*}

\begin{figure*}[t]
    \centering
    \includegraphics[width=0.9\linewidth]{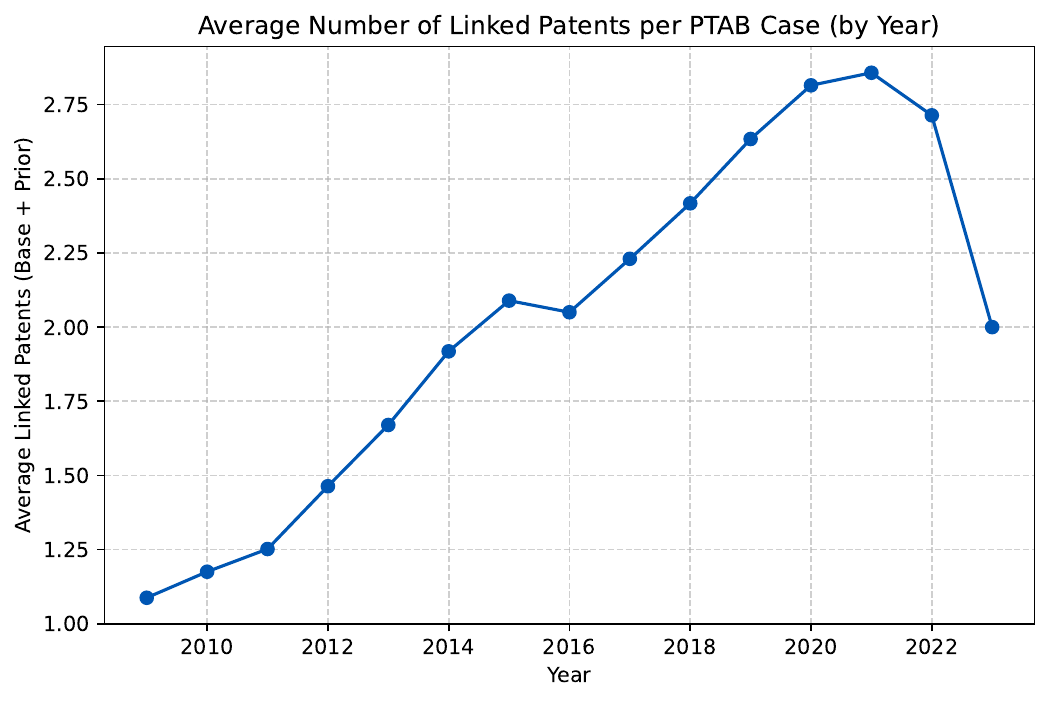}
    \caption{Average number of linked patents per PTAB case by year.}
    \label{fig:linked_case_year}
\end{figure*}
% -------

% -------------------------- Overall

\begin{table*}[t]
  \centering
  \scriptsize
  \begin{tabular}{lcccccccc}
    \toprule
    \textbf{Model} 
      & \textbf{Exact Match} 
      & \textbf{Micro-P} 
      & \textbf{Micro-R} 
      & \textbf{Micro-F1} 
      & \textbf{Macro-P} 
      & \textbf{Macro-R} 
      & \textbf{Macro-F1} 
      & \textbf{HL} \\
    \midrule

    % ---------- Split (Base) ----------
    \multicolumn{9}{c}{\textbf{Split (Base)}} \\
    \cmidrule(lr){1-9}
    Claude-Sonnet-4 &0.5871	&0.7322	&0.8589	&0.7905	&0.5340	&0.5735	&0.5457	&0.0893  \\
    Gemini-2.5-pro  &0.5874	&0.7285	&0.8683	&0.7923	&0.6427	&0.7137	&0.6630	&0.1072  \\
    GPT-4o          &0.5751	&0.7215	&0.8633	&0.7860	&0.6284	&0.6997	&0.6519	&0.1107  \\
    GPT-o3          &0.5955	&0.7404	&0.8624	&0.7968	&0.6567	&0.6969	&0.6639	&0.1036  \\
    Solar-pro2      &0.5583	&0.7072	&0.8467	&0.7707	&0.4988	&0.5653	&0.5240	&0.0989  \\
    \addlinespace[2pt]
    LLaMA-3.1(8B)   &0.1826	&0.4512	&0.8092	&0.5793	&0.0920	&0.1530	&0.1051	&0.0659  \\
    Mistral(7B)     &0.3405	&0.5302	&0.7126	&0.6080	&0.1936	&0.2650	&0.2111	&0.0902  \\
    Qwen(8B)        &0.5561	&0.7114	&0.8489	&0.7741	&0.5006	&0.5598	&0.5251	&0.0972  \\
    T5(2B)          &0.0772	&0.2945	&0.9265	&0.4469	&0.2812	&0.9118	&0.3845	&0.5401  \\
    \midrule

    % ---------- Merge ----------
    \multicolumn{9}{c}{\textbf{Merge}} \\
    \cmidrule(lr){1-9}
    Claude-Sonnet-4 &0.5879	&0.7330	&0.8602	&0.7915	&0.5348	&0.5745	&0.5468	&0.0889  \\
    Gemini-2.5-pro  &0.5810	&0.7220	&0.8694	&0.7889	&0.6351	&0.7241	&0.6625	&0.1096  \\
    GPT-4o          &0.5516	&0.6984	&0.8726	&0.7758	&0.6039	&0.7129	&0.6422	&0.1188  \\
    GPT-o3          &0.5943	&0.7375	&0.8648	&0.7961	&0.6535	&0.7025	&0.6645	&0.1043  \\
    Solar-pro2      &0.5466	&0.6919	&0.8535	&0.7643	&0.5817	&0.6975	&0.6249	&0.1240  \\
    \addlinespace[2pt]
    LLaMA-3.1(8B)   &0.1334	&0.4408	&0.8482	&0.5801	&0.3689	&0.7003	&0.4517	&0.2892  \\
    Mistral(7B)     &0.2639	&0.4631	&0.7617	&0.5760	&0.1117	&0.2013	&0.1356	&0.0777  \\
    Qwen(8B)        &0.5322	&0.6825	&0.8660	&0.7634	&0.5732	&0.6973	&0.6255	&0.1264  \\
    T5(2B)          &0.0057	&0.2563	&0.9643	&0.4050	&0.2535	&0.9624	&0.3534	&0.6674  \\
    \midrule

    % ---------- Split+Claim ----------
    \multicolumn{9}{c}{\textbf{Split+Claim}} \\
    \cmidrule(lr){1-9}
    Claude-Sonnet-4 &0.5869	&0.7339	&0.8589	&0.7915	&0.5342	&0.5707	&0.5443	&0.0888  \\
    Gemini-2.5-pro  &0.5911	&0.7334	&0.8690	&0.7955	&0.6475	&0.7062	&0.6632	&0.1052  \\
    GPT-4o          &0.5658	&0.7077	&0.8759	&0.7828	&0.6155	&0.7127	&0.6492	&0.1144  \\
    GPT-o3          &0.5946	&0.7393	&0.8639	&0.7967	&0.6550	&0.6991	&0.6639	&0.1038  \\
    Solar-pro2      &0.5355	&0.6808	&0.8589	&0.7596	&0.5736	&0.7066	&0.6225	&0.1281  \\
    \addlinespace[2pt]
    LLaMA-3.1(8B)   &0.1785	&0.4587	&0.8377	&0.5928	&0.3477	&0.6530	&0.4360	&0.2710  \\
    Mistral(7B)     &0.4200	&0.5964	&0.7820	&0.6767	&0.2439	&0.3113	&0.2662	&0.0880  \\
    Qwen(8B)        &0.5631	&0.7229	&0.8426	&0.7782	&0.6204	&0.6599	&0.6353	&0.1131  \\
    T5(2B)          &0.0155	&0.3048	&0.8931	&0.4545	&0.0018	&0.0052	&0.0024	&0.0030  \\
    \bottomrule
  \end{tabular}
  \caption{Results for the Issue Type classification task with 8 evaluation metrics. Exact Match, Micro-P (Micro-Precision), Micro-R (Macro-Recall), Micro-F1 (Micro-F1), Macro-P (Macro-Precision), Macro-R (Macro-Recall), Macro-F1 (Macro-F1) and HL (Hamming Loss) are reported.}
  \label{tab:issue-result-overall}
\end{table*}

\begin{table*}[t]
  \centering
  \scriptsize
  \begin{tabular}{lcccccccc}
    \toprule
    \textbf{Model} 
      & \textbf{Exact Match} 
      & \textbf{Micro-P} 
      & \textbf{Micro-R} 
      & \textbf{Micro-F1} 
      & \textbf{Macro-P} 
      & \textbf{Macro-R} 
      & \textbf{Macro-F1} 
      & \textbf{HL} \\
    \midrule

    % ---------- Split (Base) ----------
    \multicolumn{9}{c}{\textbf{Split (Base)}} \\
    \cmidrule(lr){1-9}
    Claude-Sonnet-4 &0.4945	&0.6038	&0.4956	&0.5444	&0.2499	&0.3503	&0.2397	&0.1012  \\
    Gemini-2.5-pro  &0.5906	&0.8158	&0.6003	&0.6916	&0.2549	&0.4277	&0.2665	&0.0725  \\
    GPT-4o          &0.6314	&0.7004	&0.6102	&0.6522	&0.3177	&0.3509	&0.2589	&0.0882  \\
    GPT-o3          &0.5302	&0.6831	&0.5736	&0.6236	&0.2787	&0.2504	&0.1940	&0.0603  \\
    Solar-pro2      &0.4293	&0.5825	&0.6279	&0.6179	&0.1054	&0.2274	&0.1014	&0.0584  \\
    \addlinespace[2pt]
    LLaMA-3.1(8B)   &0.0000	&0.0934	&0.1801	&0.1230	&0.1359	&0.3945	&0.0843	&0.3132  \\
    Mistral(7B)     &0.0028	&0.2043	&0.4263	&0.2762	&0.0100	&0.0300	&0.0075	&0.0211  \\
    Qwen(8B)        &0.1542	&0.1899	&0.2039	&0.1966	&0.1860	&0.4106	&0.1420	&0.2258  \\
    T5(2B)          &0.0064	&0.1508	&0.3548	&0.2116	&0.0030	&0.0079	&0.0026	&0.0064  \\
    \midrule

    % ---------- Merge ----------
    \multicolumn{9}{c}{\textbf{Merge}} \\
    \cmidrule(lr){1-9}
    Claude-Sonnet-4 &0.7761	&0.8924	&0.7304	&0.8033	&0.2105	&0.2919	&0.2128	&0.0364  \\
    Gemini-2.5-pro  &0.6323	&0.9148	&0.6194	&0.7387	&0.3551	&0.4168	&0.3062	&0.0594  \\
    GPT-4o          &0.6032	&0.6525	&0.5868	&0.6179	&0.2419	&0.4041	&0.2486	&0.0984  \\
    GPT-o3          &0.6459	&0.8436	&0.6503	&0.7344	&0.2732	&0.2705	&0.2160	&0.0441  \\
    Solar-pro2      &0.2531	&0.4928	&0.6284	&0.5524	&0.0628	&0.1502	&0.0620	&0.0460  \\
    \addlinespace[2pt]
    LLaMA-3.1(8B)   &0.0000	&0.1169	&0.2685	&0.1629	&0.1218	&0.3772	&0.0882	&0.3061  \\
    Mistral(7B)     &0.0028	&0.1984	&0.4372	&0.2729	&0.0050	&0.0146	&0.0038	&0.0112  \\
    Qwen(8B)        &0.4266	&0.4641	&0.4427	&0.4531	&0.1960	&0.3699	&0.1897	&0.1448  \\
    T5(2B)          &0.0026	&0.1105	&0.4283	&0.1757	&0.0035	&0.0117	&0.0032	&0.0099  \\
    \midrule

    % ---------- Split+Claim ----------
    \multicolumn{9}{c}{\textbf{Split+Claim}} \\
    \cmidrule(lr){1-9}
    Claude-Sonnet-4 &0.2026	&0.2920	&0.2402	&0.2636	&0.1838	&0.2837	&0.1530	&0.1364  \\
    Gemini-2.5-pro  &0.4913	&0.6261	&0.5394	&0.5795	&0.2122	&0.4493	&0.2201	&0.1061  \\
    GPT-4o          &0.0035	&0.1206	&0.1760	&0.1431	&0.1806	&0.4817	&0.1425	&0.2856  \\
    GPT-o3          &0.2477	&0.4011	&0.4396	&0.4194	&0.2444	&0.2991	&0.2109	&0.1060  \\
    Solar-pro2      &0.0041	&0.1596	&0.2011	&0.1780	&0.0732	&0.2122	&0.0485	&0.1133  \\
    \addlinespace[2pt]
    LLaMA-3.1(8B)   &0.0001	&0.1408	&0.3171	&0.1950	&0.1296	&0.3130	&0.0923	&0.2904  \\
    Mistral(7B)     &0.0003	&0.1154	&0.2627	&0.1603	&0.0070	&0.0197	&0.0044	&0.0185  \\
    Qwen(8B)        &0.0134	&0.0544	&0.0606	&0.0574	&0.1917	&0.3804	&0.1136	&0.2700  \\
    T5(2B)          &0.0009	&0.0912	&0.3431	&0.1442	&0.0051	&0.0248	&0.0037	&0.0206  \\
    \bottomrule
  \end{tabular}
  \caption{Results for the Board Authorities classification task with 8 evaluation metrics. Exact Match, Micro-P (Micro-Precision), Micro-R (Macro-Recall), Micro-F1 (Micro-F1), Macro-P (Macro-Precision), Macro-R (Macro-Recall), Macro-F1 (Macro-F1) and HL (Hamming Loss) are reported.}
  \label{tab:board-result-overall}
\end{table*}

\begin{table*}[t]
  \centering
  \scriptsize
  \begin{tabular}{lccccccc}
    \toprule
    \textbf{Model}
      & \textbf{Acc}
      & \textbf{Balanced Acc}
      & \textbf{Macro-P}
      & \textbf{Macro-R}
      & \textbf{Macro-F1}
      & \textbf{Micro-F1}
      & \textbf{Weighted-F1} \\
    \midrule

    % ---------- Split (Base) ----------
    \multicolumn{8}{c}{\textbf{Split (Base)}} \\
    \cmidrule(lr){1-8}
    Claude-Sonnet-4 &0.5658	&0.1681	&0.1767	&0.1569	&0.1296	&0.5658	&0.4854  \\
    Gemini-2.5-pro  &0.5050	&0.1765	&0.2473	&0.1647	&0.1635	&0.5050	&0.4982  \\
    GPT-4o          &0.4924	&0.1327	&0.0944	&0.1283	&0.0997	&0.4924	&0.4709  \\
    GPT-o3          &0.5918	&0.1519	&0.3295	&0.1519	&0.1639	&0.5918	&0.5541  \\
    Solar-pro2      &0.5369	&0.1225	&0.1509	&0.1143	&0.0779	&0.5369	&0.3923  \\
    \addlinespace[2pt]
    LLaMA-3.1(8B)   &0.4364	&0.0927	&0.0841	&0.0927	&0.0767	&0.4364	&0.4006  \\
    Mistral(7B)     &0.1241	&0.0603	&0.0461	&0.0422	&0.0251	&0.1241	&0.1284  \\
    Qwen(8B)        &0.4793	&0.1106	&0.1057	&0.1032	&0.0977	&0.4793	&0.4457  \\
    T5(2B)          &0.0419	&0.0917	&0.0501	&0.0583	&0.0142	&0.0419	&0.0617  \\
    \midrule

    % ---------- Merge ----------
    \multicolumn{8}{c}{\textbf{Merge}} \\
    \cmidrule(lr){1-8}
    Claude-Sonnet-4 &0.5590	&0.1614	&0.1872	&0.1509	&0.1129	&0.5590	&0.4320  \\
    Gemini-2.5-pro  &0.5114	&0.1925	&0.1661	&0.1685	&0.1443	&0.5114	&0.5036  \\
    GPT-4o          &0.4592	&0.1257	&0.1381	&0.1173	&0.0912	&0.4592	&0.4353  \\
    GPT-o3          &0.6086	&0.1580	&0.3244	&0.1580	&0.1683	&0.6086	&0.5682  \\
    Solar-pro2      &0.5420	&0.1248	&0.1790	&0.1164	&0.0804	&0.5420	&0.3932  \\
    \addlinespace[2pt]
    LLaMA-3.1(8B)   &0.5036	&0.0650	&0.0536	&0.5036	&0.0696	&0.3971	&0.0676  \\
    Mistral(7B)     &0.1265	&0.0364	&0.0229	&0.1265	&0.0572	&0.1249	&0.0407  \\
    Qwen(8B)        &0.4266	&0.1096	&0.0707	&0.0768	&0.0698	&0.4266	&0.4264  \\
    T5(2B)          &0.0191	&0.0463	&0.0092	&0.0191	&0.0794	&0.0270	&0.0437  \\
    \midrule

    % ---------- Split+Claim ----------
    \multicolumn{8}{c}{\textbf{Split+Claim}} \\
    \cmidrule(lr){1-8}
    Claude-Sonnet-4 &0.5620	&0.1616	&0.1725	&0.1509	&0.1272	&0.5620	&0.4842  \\
    Gemini-2.5-pro  &0.4908	&0.1518	&0.1832	&0.1417	&0.1433	&0.4908	&0.4854  \\
    GPT-4o          &0.3804	&0.1275	&0.0944	&0.1190	&0.0892	&0.3804	&0.3581  \\
    GPT-o3          &0.5884	&0.1610	&0.3241	&0.1610	&0.1692	&0.5884	&0.5538  \\
    Solar-pro2      &0.5373	&0.0762	&0.0993	&0.0762	&0.0608	&0.5373	&0.3966  \\
    \addlinespace[2pt]
    LLaMA-3.1(8B)   &0.4125	&0.0664	&0.0830	&0.0664	&0.0642	&0.4125	&0.3938  \\
    Mistral(7B)     &0.1209	&0.0536	&0.0533	&0.0417	&0.0295	&0.1209	&0.1205  \\
    Qwen(8B)        &0.4368	&0.0872	&0.0831	&0.0814	&0.0794	&0.4368	&0.4364  \\
    T5(2B)          &0.0225	&0.1699	&0.1655	&0.1322	&0.0436	&0.0225	&0.0168  \\
    \bottomrule
  \end{tabular}
  \caption{Results for the Subdecision (Fine-grained) classification task with 7 evaluation metrics. Acc (Accuracy), Balanced Acc (Balanced Accuracy), Macro-P (Macro-Precision), Macro-R (Macro-Recall), Macro-F1 (Macro-F1), Micro-F1 (Micro-F1), and Weighted-F1 are reported. In single-label multiclass classification, Accuracy and Micro-F1 coincide because both measure the proportion of correctly classified samples.}
  \label{tab:subdecision-fine-result-overall}
\end{table*}

\begin{table*}[t]
  \centering
  \scriptsize
  \begin{tabular}{lccccccc}
    \toprule
    \textbf{Model}
      & \textbf{Acc}
      & \textbf{Balanced Acc}
      & \textbf{Macro-P}
      & \textbf{Macro-R}
      & \textbf{Macro-F1}
      & \textbf{Micro-F1}
      & \textbf{Weighted-F1} \\
    \midrule

    % ---------- Split (Base) ----------
    \multicolumn{8}{c}{\textbf{Split (Base)}} \\
    \cmidrule(lr){1-8}
    Claude-Sonnet-4 &0.5652	&0.2108	&0.2865	&0.2105	&0.2116	&0.5625	&0.4900  \\
    Gemini-2.5-pro  &0.5063	&0.2270	&0.3351	&0.2270	&0.2366	&0.5063	&0.4927  \\
    GPT-4o          &0.5045	&0.1988	&0.2350	&0.1988	&0.2037	&0.5045	&0.4863  \\
    GPT-o3          &0.5863	&0.2099	&0.3802	&0.2099	&0.2126	&0.5863	&0.5511  \\
    Solar-pro2      &0.5389	&0.1621	&0.2303	&0.1621	&0.1356	&0.5389	&0.3929  \\
    \addlinespace[2pt]
    LLaMA-3.1(8B)   &0.4764	&0.1635	&0.1770	&0.1635	&0.1551	&0.4764	&0.4024  \\
    Mistral(7B)     &0.0726	&0.1590	&0.1725	&0.1590	&0.0758	&0.0726	&0.0994  \\
    Qwen(8B)        &0.4733	&0.1739	&0.2298	&0.1739	&0.1692	&0.4733	&0.4404  \\
    T5(2B)          &0.0254	&0.2177	&0.1446	&0.2177	&0.0499	&0.0254	&0.0146  \\
    \midrule

    % ---------- Merge ----------
    \multicolumn{8}{c}{\textbf{Merge}} \\
    \cmidrule(lr){1-8}
    Claude-Sonnet-4 &0.5607	&0.1952	&0.2872	&0.1952	&0.1788	&0.5607	&0.4456  \\
    Gemini-2.5-pro  &0.5119	&0.2390	&0.2771	&0.2390	&0.2381	&0.5119	&0.5001  \\
    GPT-4o          &0.4972	&0.1794	&0.2635	&0.1794	&0.1820	&0.4972	&0.4638  \\
    GPT-o3          &0.6020	&0.2101	&0.3814	&0.2101	&0.2125	&0.6020	&0.5631  \\
    Solar-pro2      &0.5423	&0.1631	&0.2598	&0.1631	&0.1390	&0.5423	&0.3967  \\
    \addlinespace[2pt]
    LLaMA-3.1(8B)   &0.5229	&0.1515	&0.1908	&0.1515	&0.1253	&0.5229	&0.3922  \\
    Mistral(7B)     &0.0823	&0.1552	&0.1685	&0.1552	&0.0821	&0.0823	&0.1168  \\
    Qwen(8B)        &0.4163	&0.1760	&0.2219	&0.1760	&0.1761	&0.4163	&0.4223  \\
    T5(2B)          &0.0234	&0.2238	&0.1593	&0.2238	&0.0446	&0.0234	&0.0092  \\
    \midrule

    % ---------- Split+Claim ----------
    \multicolumn{8}{c}{\textbf{Split+Claim}} \\
    \cmidrule(lr){1-8}
    Claude-Sonnet-4 &0.5639	&0.2011	&0.2646	&0.2011	&0.2018	&0.5637	&0.4889  \\
    Gemini-2.5-pro  &0.4915	&0.2142	&0.3409	&0.2142	&0.2111	&0.4915	&0.4840  \\
    GPT-4o          &0.3046	&0.1633	&0.1982	&0.1633	&0.1206	&0.3046	&0.2027  \\
    GPT-o3          &0.5783	&0.2099	&0.5012	&0.2099	&0.2068	&0.5783	&0.5426  \\
    Solar-pro2      &0.5364	&0.1514	&0.1819	&0.1514	&0.1210	&0.5364	&0.3977  \\
    \addlinespace[2pt]
    LLaMA-3.1(8B)   &0.4741	&0.1447	&0.1505	&0.1447	&0.1259	&0.4741	&0.3909  \\
    Mistral(7B)     &0.0587	&0.1568	&0.2767	&0.1568	&0.0549	&0.0587	&0.0721  \\
    Qwen(8B)        &0.4605	&0.1660	&0.2083	&0.1660	&0.1655	&0.4605	&0.4439  \\
    T5(2B)          &0.0136	&0.0440	&0.0376	&0.0246	&0.0053	&0.0136	&0.0142  \\
    \bottomrule
  \end{tabular}
  \caption{Results for the Subdecision (Coarse-grained) classification task with 7 evaluation metrics. Acc (Accuracy), Balanced Acc (Balanced Accuracy), Macro-P (Macro-Precision), Macro-R (Macro-Recall), Macro-F1 (Macro-F1), Micro-F1 (Micro-F1), and Weighted-F1 are reported. In single-label multiclass classification, Accuracy and Micro-F1 coincide because both measure the proportion of correctly classified samples.}
  \label{tab:subdecision-coarse-result-overall}
\end{table*}

% -------------------------- T5 recall

\begin{figure*}[t]
    \centering
    \includegraphics[width=1\linewidth]{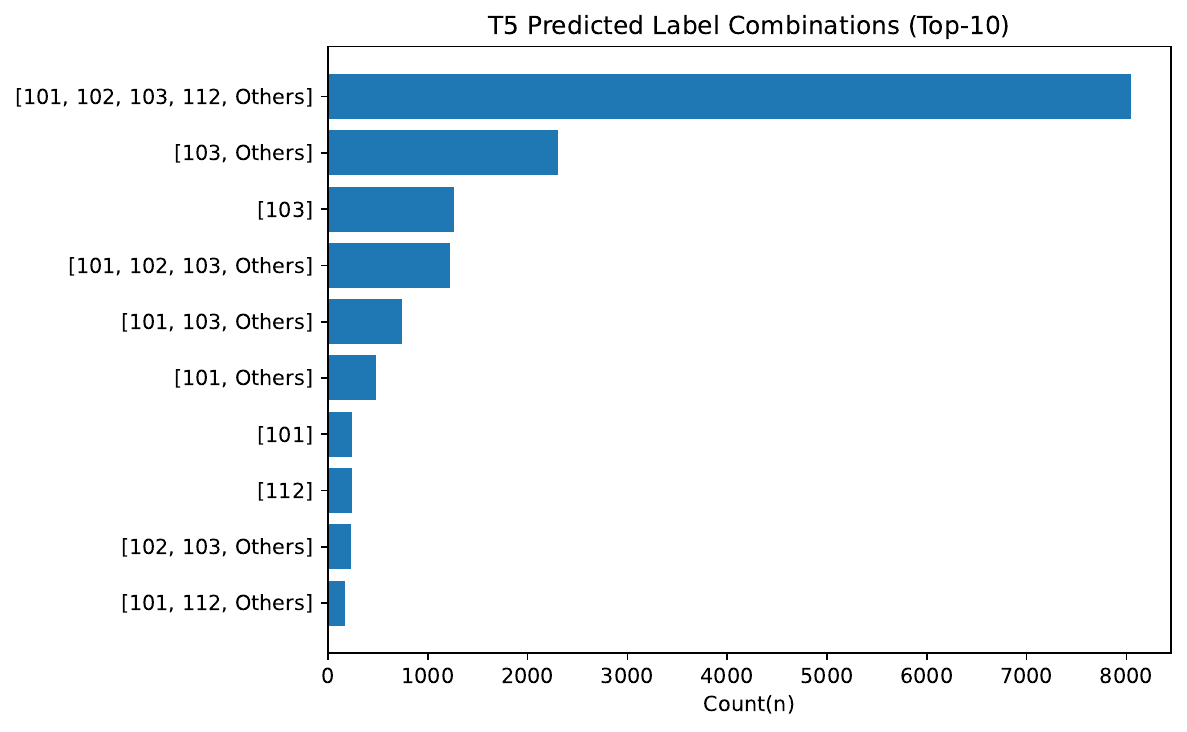}
    \caption{Top-10 predicted IssueType label combinations by T5 under Split (Base).}
    \label{fig:t5-base}
\end{figure*}
\begin{figure*}[t]
    \centering
    \includegraphics[width=1\linewidth]{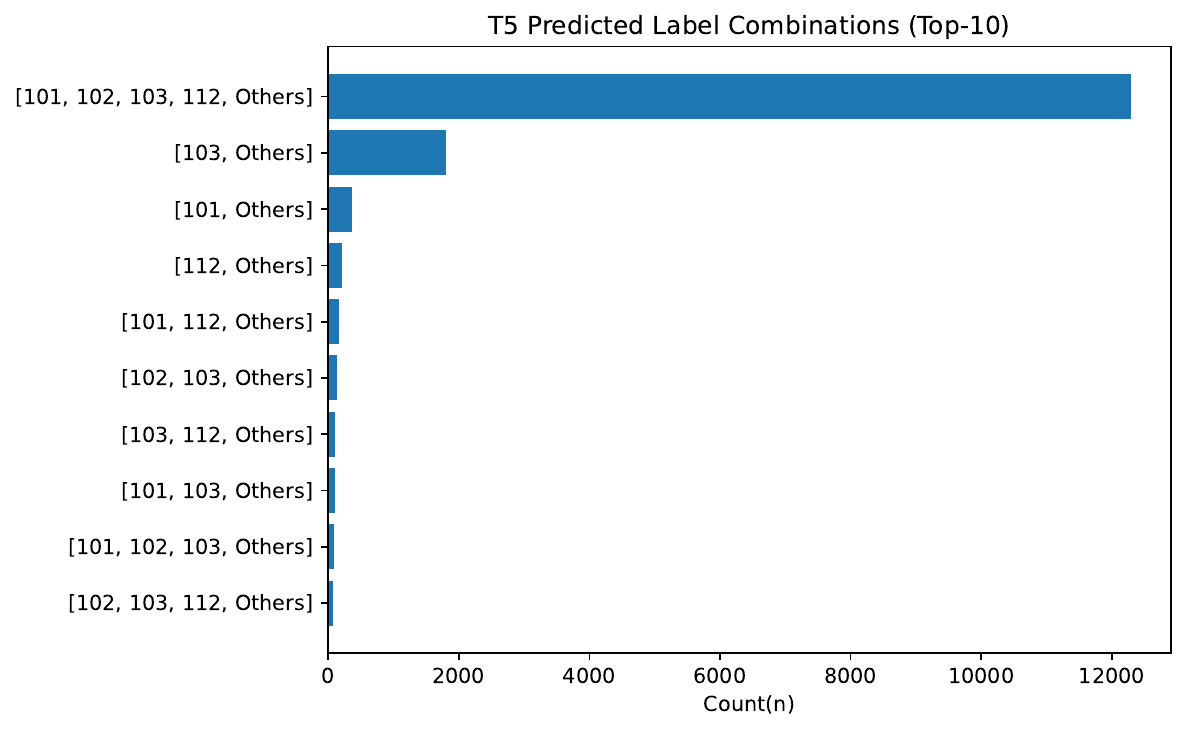}
    \caption{Top-10 predicted IssueType label combinations by T5 under Merge.}
    \label{fig:t5-merge}
\end{figure*}
\begin{figure*}[t]
    \centering
    \includegraphics[width=1\linewidth]{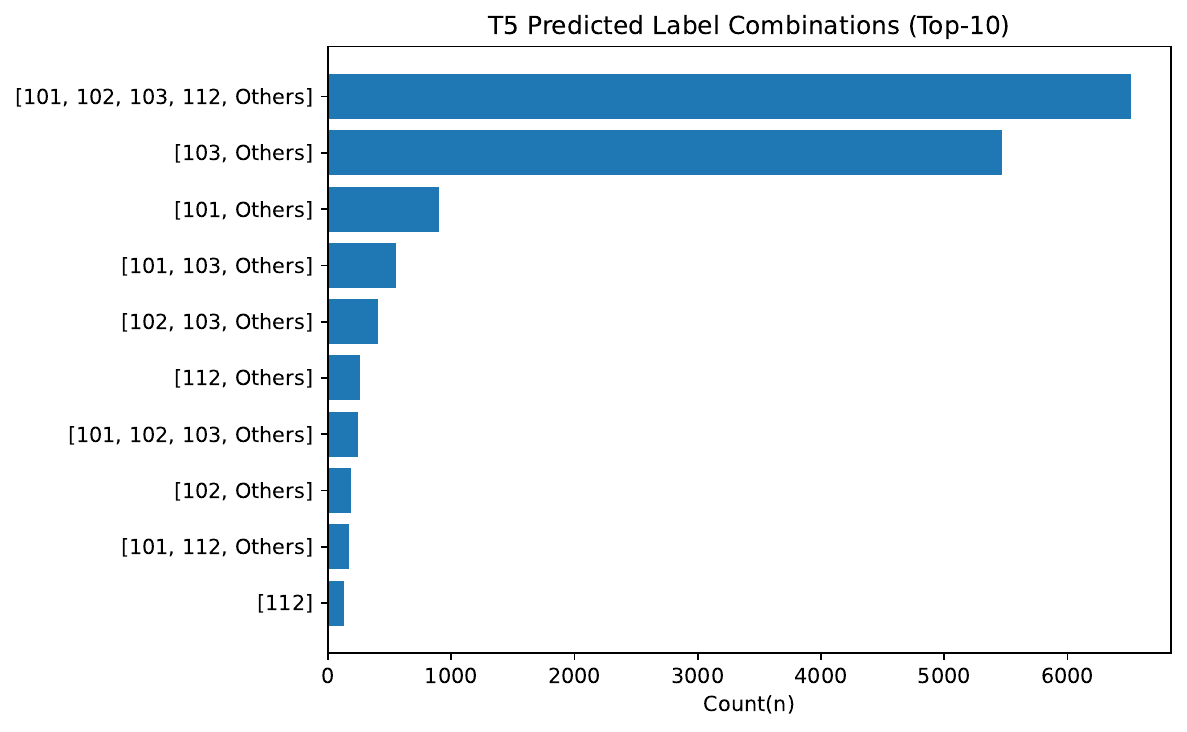}
    \caption{Top-10 predicted IssueType label combinations by T5 under Split+Claim.}
    \label{fig:t5-claim}
\end{figure*}

% -------------------------- Labels

\begin{table*}[t]
  \centering
  \scriptsize
  \begin{tabularx}{\columnwidth}{l X}
    \toprule
    \textbf{Label} & \textbf{Definition} \\
    \midrule
    101 & Patent eligibility (Subject-matter eligibility) \\
    102 & Novelty \\
    103 & Non-obviousness \\
    112 & Specification requirements (Written description / Enablement / Definiteness) \\
    Others & All other issues (e.g., OTDP, priority, new matter, reissue, design) \\
    \bottomrule
  \end{tabularx}
  \caption{Labels used in the Issue Type classification task and their definitions. The dictionary was also provided within the classification prompt so that the LLM could reference these descriptions while reasoning about applicable statutory issues.}
  \label{tab:appendix-issue-def}
\end{table*}

\begin{table*}[t]
  \centering
  \scriptsize
  \begin{tabularx}{\textwidth}{l X}
    \toprule
    \textbf{Label} & \textbf{Definition} \\
    \midrule
    37 CFR 41.50   & General framework for PTAB decisions/actions in ex parte appeals (affirm/reverse/remand, new ground, additional briefing, time extensions). \\
    37 CFR 41.50(a) & Merits decision on appeal (affirm/reverse/remand) and post-decision options. \\
    37 CFR 41.50(b) & Board-designated New Ground of Rejection (non-final for judicial review); appellant may request rehearing or reopen prosecution. \\
    37 CFR 41.50(c) & Procedure to address an undesignated new ground via rehearing request. \\
    37 CFR 41.50(d) & Authority to order additional briefing/information; non-compliance may lead to dismissal. \\
    37 CFR 41.50(f) & Rules for extensions of time for replies in ex parte appeals. \\
    37 CFR 1.131   & Pre-AIA affidavit/declaration of prior invention (swear behind) to overcome prior art. \\
    37 CFR 1.132   & Affidavits/declarations traversing rejections or objections (e.g., objective evidence, secondary considerations). \\
    35 USC 251     & Reissue of defective patents (broadening/narrowing; correction of error). \\
    35 USC 161     & Plant patent requirements (asexual reproduction, cultivar/variety). \\
    \bottomrule
  \end{tabularx}
  \caption{Labels used in the Board Authorities classification task and their definitions. This dictionary was also embedded in the classification prompt, so that the LLM could reference these descriptions while reasoning and assigning labels.}
  \label{tab:appendix-board-authorities-def}
\end{table*}

\begin{table*}[t]
  \centering
  \scriptsize
  \begin{tabularx}{\textwidth}{r l X} % ← r: Index, l: Label, X: Variants
    \toprule
    \textbf{ID} & \textbf{Label} & \textbf{Variants / Mappings} \\
    \midrule
    1 & Affirmed & affirmed \\
    \midrule

    \multirow{3}{*}{2} & \multirow{3}{*}{Affirmed with New Ground of Rejection}
      & affirmed with new ground of rejection \\
      & & affirmed with new ground(s) of rejection \\
      & & affirmed w/ new ground(s) of rejection \\
    \midrule

    \multirow{8}{*}{3} & \multirow{8}{*}{Affirmed-in-Part}
      & affirmed-in-part \\
      & & affirmed in part \\
      & & affirmed-in part \\
      & & affirmed/reversed in part \\
      & & reversed/affirmed in part \\
      & & reversed in-part \\
      & & reversed in part \\
      & & reversed-in part \\
    \midrule

    \multirow{2}{*}{4} & \multirow{2}{*}{Affirmed-in-Part and Remanded}
      & affirmed-in-part and remanded \\
      & & affirmed-in-part and remanded with new ground of rejection \\
    \midrule

    \multirow{3}{*}{5} & \multirow{3}{*}{Affirmed-in-Part with New Ground of Rejection}
      & affirmed-in-part with new ground of rejection \\
      & & affirmed-in-part with new ground(s) of rejection \\
      & & affirmed-in-part w/ new ground(s) of rejection \\
    \midrule

    6 & Reversed & reversed \\
    \midrule

    \multirow{3}{*}{7} & \multirow{3}{*}{Reversed with New Ground of Rejection}
      & reversed with new ground of rejection \\
      & & reversed with new ground(s) of rejection \\
      & & reversed w/ new ground(s) of rejection \\
    \midrule

    8 & Reexam affirmed & reexam affirmed \\
    \midrule
    9 & Reexam Affirmed-in-part & reexam affirmed-in-part \\
    \midrule
    10 & Reexam Affirmed-in-part with New Ground of Rejection & reexam affirmed-in-part with new ground of rejection \\
    \midrule
    11 & Reexam reversed & reexam reversed \\
    \midrule

    12 & Inter Partes Reexam Affirmed & inter partes reexam affirmed \\
    \midrule
    13 & Inter Partes Reexam Affirmed-in-part & inter partes reexam affirmed-in-part \\
    \midrule
    14 & Inter Partes Reexam Reversed & inter partes reexam reversed \\
    \midrule
    15 & Inter Partes Reexam New Ground of Rejection & inter partes reexam new ground of rejection \\
    \midrule
    16 & Inter partes reexam rehearing decision is a new decision & inter partes reexam rehearing decision is a new decision \\
    \midrule

    17 & Affirmed-in-Part and Remanded with New Ground of Rejection & affirmed-in-part and remanded with new ground of rejection \\
    \midrule
    18 & Reversed and Remanded & reversed and remanded \\
    \midrule

    \multirow{8}{*}{19} & \multirow{8}{*}{Vacated}
      & vacated \\
      & & vacated with new ground of rejection \\
      & & vacated-in-part with new ground of rejection \\
      & & vacated/remanded \\
      & & vacated and remanded \\
      & & vacatur \\
      & & vacated in part \\
      & & vacate and remand \\
    \midrule

    \multirow{6}{*}{20} & \multirow{6}{*}{Granted}
      & granted \\
      & & granted (petitioner) \\
      & & granted (patent owner) \\
      & & granted-in-part \\
      & & granted-in-part (petitioner) \\
      & & granted-in-part (patent owner) \\
    \midrule

    \multirow{3}{*}{21} & \multirow{3}{*}{Denied}
      & denied \\
      & & denied (petitioner) \\
      & & denied (patent owner) \\
    \midrule

    \multirow{4}{*}{22} & \multirow{4}{*}{Rehearing Decision - Granted}
      & rehearing decision - granted \\
      & & Rehearing Decision \verb|â| Grante \\
      & & rehearing decision - granted \\
      & & rehearing decision-granted \\
    \midrule

    23 & Reexam rehearing decision final and appealable & reexam rehearing decision final and appealable \\
    \bottomrule
  \end{tabularx}
  \caption{Normalized subdecision fine categories (excluding \textbf{Others}) and their variants. Each variant was normalized by converting raw labels to lowercase and stripping leading/trailing whitespace before mapping them to a canonical label. The canonical labels are further incorporated into the classification prompt, enabling the LLM to consult these standardized categories during subdecision reasoning.}
  \label{tab:subdecision-fine-mapping-main}
\end{table*}

\begin{table*}[t]
  \centering
  \scriptsize
  \begin{tabularx}{\textwidth}{l X}
    \toprule
    \textbf{Label} & \textbf{Variants / Mappings} \\
    \midrule
    \multirow{45}{*}{Others} % 행 개수는 실제 줄 수에 맞게 조정
      & dismissed \\
      & dismissal \\
      & voluntarily dismissed \\
      & dismissed before institution \\
      & dismissed after institution \\
      & decision on rehearing \\
      & decision on petition \\
      & rehearing decision \\
      & Rehearing Decision \verb|â| Granted w/ New Ground of Rejection \\
      & rehearing decision - granted with new ground of rejection \\
      & Rehearing Decision \verb|â| Denied \\
      & rehearing decision - denied \\
      & Rehearing Decision \verb|â| Denied w/ New Ground of Rejection \\
      & rehearing decision - denied with new ground of rejection \\
      & Rehearing Decision \verb|â| Granted-in-Part \\
      & rehearing decision - granted-in-part \\
      & remand \\
      & administrative remand \\
      & affirmed and remanded \\
      & reverse and remanded with new ground of rejection \\
      & panel remand \\
      & panel remand with new ground of rejection \\
      & remanded-in part \\
      & institution granted \\
      & institution granted (joined) \\
      & institution denied \\
      & decision on petition - denied \\
      & settlement \\
      & settlement before institution \\
      & settlement after institution \\
      & settled before institution \\
      & settled after institution \\
      & termination \\
      & terminated \\
      & termination before institution \\
      & termination after institution \\
      & request for adverse judgment before institution \\
      & request for adverse judgment after institution \\
      & institution-rehearing hybrid \\
      & po rehearing request granted on institution decision granted (trial denied) \\
      & petitioner's rehearing request granted on institution decision denied (reinstituted) \\
      & final decision \\
      & final written decision \\
      & final written decision on cafc remand \\
      & subsequent final written decision after rehearing \\
      & subsequent decision \\
      & judgment \\
      & adverse judgment \\
      & decision on motion \\
      & order \\
      & order on rehearing \\
    \bottomrule
  \end{tabularx}
  \caption{Variants mapped to Others. The Others category serves as a residual class, collecting normalized raw labels that did not align with any of the explicit subdecision fine categories.}
  \label{tab:subdecision-fine-mapping-others}
\end{table*}

\begin{table*}[t]
  \centering
  \scriptsize
  \begin{tabularx}{\textwidth}{r l L} % r: ID, l: Label, L: Variants
    \toprule
    \textbf{ID} & \textbf{Label} & \textbf{Variants / Mappings} \\
    \midrule

    1 & Affirmed & affirmed \\
    \midrule

    \multirow{3}{*}{2} & \multirow{3}{*}{Affirmed with New Ground of Rejection}
      & affirmed with new ground of rejection \\
      & & affirmed with new ground(s) of rejection \\
      & & affirmed w/ new ground(s) of rejection \\
    \midrule

    \multirow{8}{*}{3} & \multirow{8}{*}{Affirmed-in-Part}
      & affirmed-in-part \\
      & & affirmed in part \\
      & & affirmed-in part \\
      & & affirmed/reversed in part \\
      & & reversed/affirmed in part \\
      & & reversed in-part \\
      & & reversed in part \\
      & & reversed-in part \\
    \midrule

    \multirow{3}{*}{4} & \multirow{3}{*}{Affirmed-in-Part with New Ground of Rejection}
      & affirmed-in-part with new ground of rejection \\
      & & affirmed-in-part with new ground(s) of rejection \\
      & & affirmed-in-part w/ new ground(s) of rejection \\
    \midrule

    5 & Reversed & reversed \\
    \midrule

    \multirow{3}{*}{6} & \multirow{3}{*}{Reversed with New Ground of Rejection}
      & reversed with new ground of rejection \\
      & & reversed with new ground(s) of rejection \\
      & & reversed w/ new ground(s) of rejection \\
    \bottomrule
  \end{tabularx}
  \caption{Normalized subdecision coarse categories (excluding \textbf{Others}) and their variants. Each variant was normalized by converting raw labels to lowercase and stripping leading/trailing whitespace before mapping them to a canonical category. The canonical labels are further incorporated into the classification prompt, enabling the LLM to consult these standardized categories during subdecision reasoning.}
  \label{tab:subdecision-coarse-mapping-main}
\end{table*}

\begin{table*}[t]
  \centering
  \scriptsize
  \begin{tabularx}{\textwidth}{l X}
    \toprule
    \textbf{Label} & \textbf{Variants / Mappings} \\
    \midrule
    \multirow{70}{*}{Others}
      & reexam affirmed \\
      & inter partes reexam affirmed \\
      & reexam affirmed-in-part \\
      & inter partes reexam affirmed-in-part \\
      & reexam affirmed-in-part with new ground of rejection \\
      & reexam reversed \\
      & inter partes reexam reversed \\
      & inter partes reexam new ground of rejection \\
      & reexam rehearing decision final and appealable \\
      & inter partes reexam rehearing decision is a new decision \\
      & granted \\
      & granted (petitioner) \\
      & granted (patent owner) \\
      & granted-in-part \\
      & granted-in-part (petitioner) \\
      & granted-in-part (patent owner) \\
      & denied \\
      & denied (petitioner) \\
      & denied (patent owner) \\
      & dismissed \\
      & dismissal \\
      & voluntarily dismissed \\
      & dismissed before institution \\
      & dismissed after institution \\
      & decision on rehearing \\
      & decision on petition \\
      & rehearing decision \\
      & Rehearing Decision \verb|â| Granted \\
      & rehearing decision - granted \\
      & rehearing decision-granted \\
      & Rehearing Decision \verb|â| Granted w/ New Ground of Rejection \\
      & rehearing decision - granted with new ground of rejection \\
      & Rehearing Decision \verb|â| Denied \\
      & rehearing decision - denied \\
      & Rehearing Decision \verb|â| Denied w/ New Ground of Rejection \\
      & rehearing decision - denied with new ground of rejection \\
      & Rehearing Decision \verb|â| Granted-in-Part \\
      & rehearing decision - granted-in-part \\
      & remand \\
      & administrative remand \\
      & affirmed-in-part and remanded \\
      & affirmed-in-part and remanded with new ground of rejection \\
      & affirmed and remanded \\
      & reversed and remanded \\
      & reverse and remanded with new ground of rejection \\
      & panel remand \\
      & panel remand with new ground of rejection \\
      & remanded-in part \\
      & vacated \\
      & vacated with new ground of rejection \\
      & vacated-in-part with new ground of rejection \\
      & vacated/remanded \\
      & vacated and remanded \\
      & vacatur \\
      & vacated in part \\
      & vacate and remand \\
      & institution granted \\
      & institution granted (joined) \\
      & institution denied \\
      & decision on petition - denied \\
      & settlement \\
      & settlement before institution \\
      & settlement after institution \\
      & settled before institution \\
      & settled after institution \\
      & termination \\
      & terminated \\
      & termination before institution \\
      & termination after institution \\
      & request for adverse judgment before institution \\
      & request for adverse judgment after institution \\
      & institution-rehearing hybrid \\
      & po rehearing request granted on institution decision granted (trial denied) \\
      & petitioner's rehearing request granted on institution decision denied (reinstituted) \\
      & final decision \\
      & final written decision \\
      & final written decision on cafc remand \\
      & subsequent final written decision after rehearing \\
      & subsequent decision \\
      & judgment \\
      & adverse judgment \\
      & decision on motion \\
      & order \\
      & order on rehearing \\
    \bottomrule
  \end{tabularx}
  \caption{Variants mapped to Others. The Others category serves as a residual class, collecting normalized raw labels that did not align with any of the explicit subdecision coarse categories.}
  \label{tab:subdecision-coarse-mapping-others}
\end{table*}

% --------------------

\begin{figure*}[t]
    \centering
    \includegraphics[width=\textwidth]{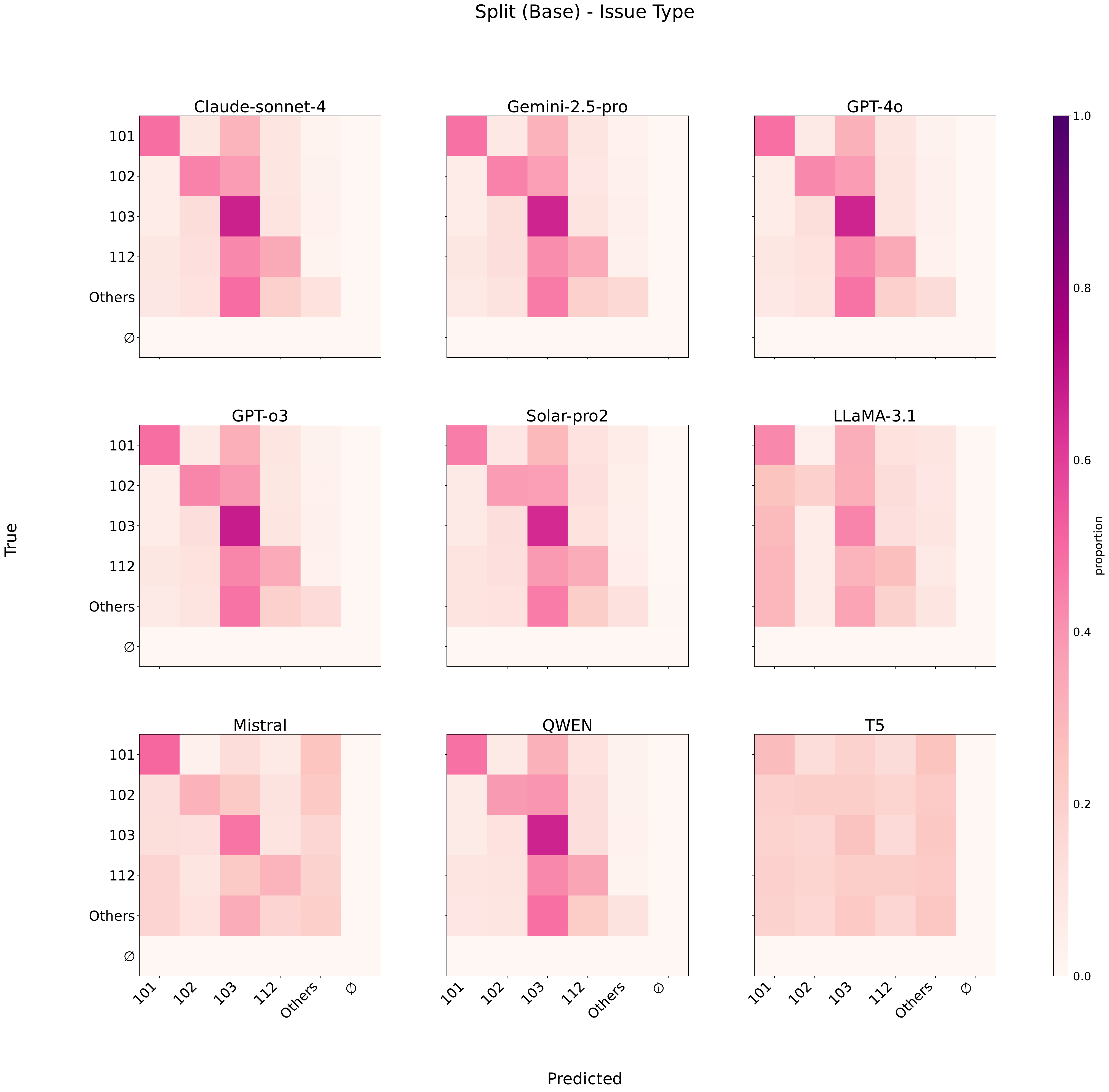}
    \caption{Heatmaps of model performance on the Issue Type classification task under the Split (Base) input setting. Each subplot visualizes the distribution of predicted versus true labels across models.}
    \label{fig:base_issue_type_all_heatmap}
\end{figure*}

\begin{figure*}[t]
    \centering
    \includegraphics[width=\textwidth]{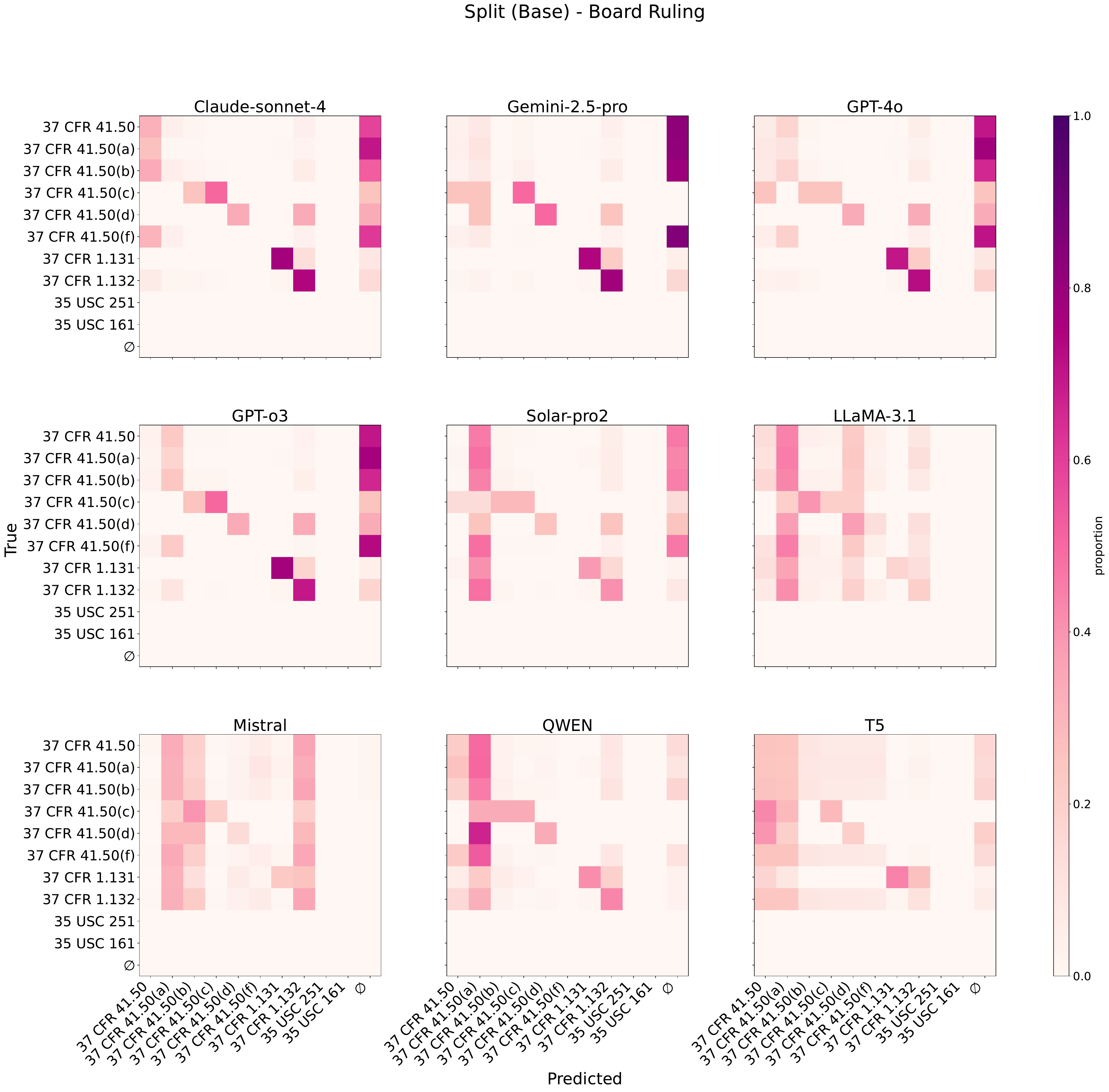}
    \caption{Heatmaps of model performance on the Board Authorities classification task under the Split (Base) input setting. Each subplot visualizes the distribution of predicted versus true labels across models.}
    \label{fig:base_board_authorities_all_heatmap}
\end{figure*}

\begin{figure*}[t]
    \centering
    \includegraphics[width=\textwidth]{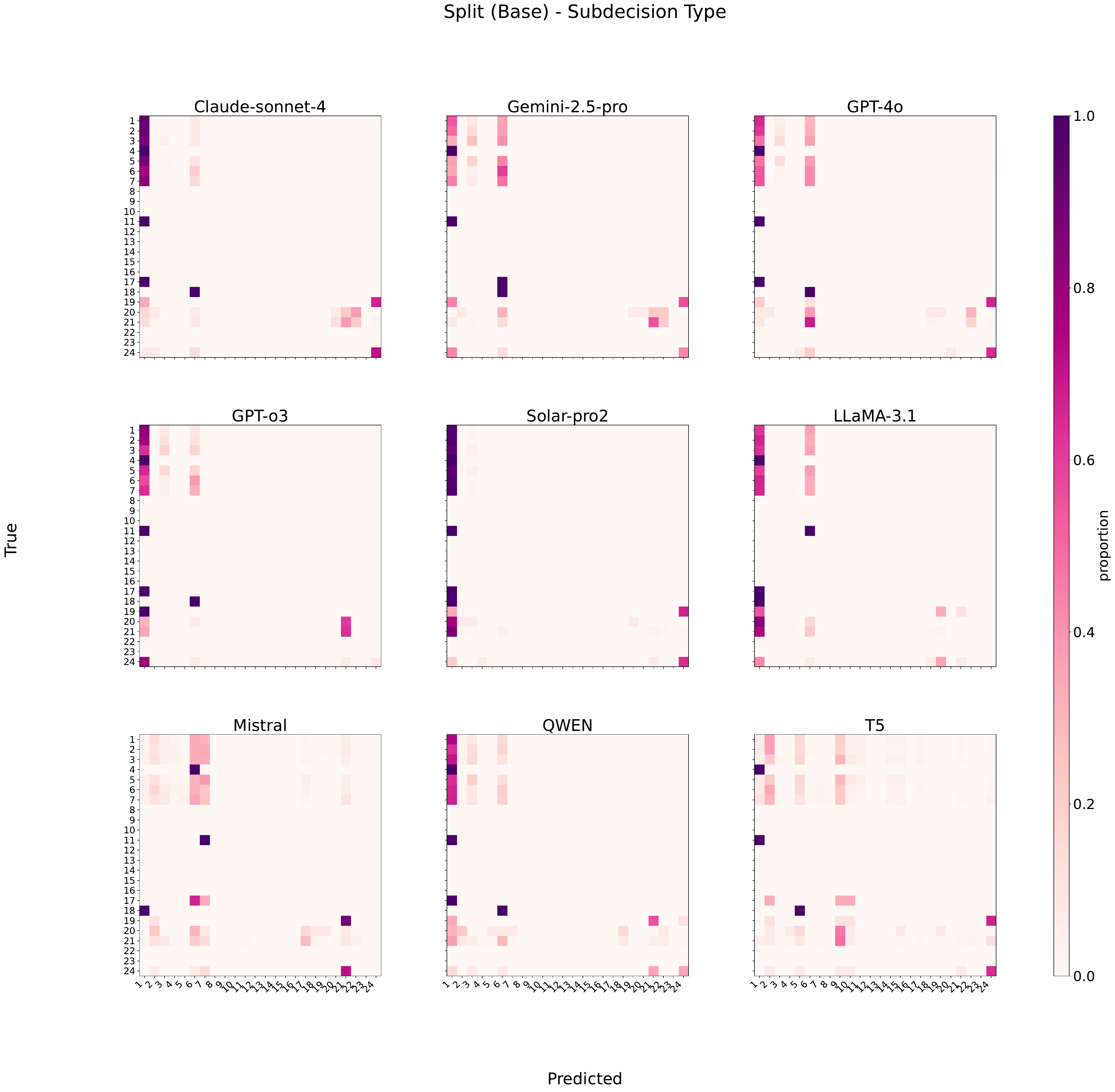}
    \caption{Heatmaps of model performance on the Subdecision (Fine-grained) classification task under the Split (Base) input setting. Each subplot visualizes the distribution of predicted versus true labels across models. The numerical indices on the axes correspond to the canonical labels defined in Table~\ref{tab:subdecision-fine-mapping-main}, where each index maps to a specific subdecision category.}
    \label{fig:base_subdecision_fine_all_heatmap}
\end{figure*}

\begin{figure*}[t]
    \centering
    \includegraphics[width=\textwidth]{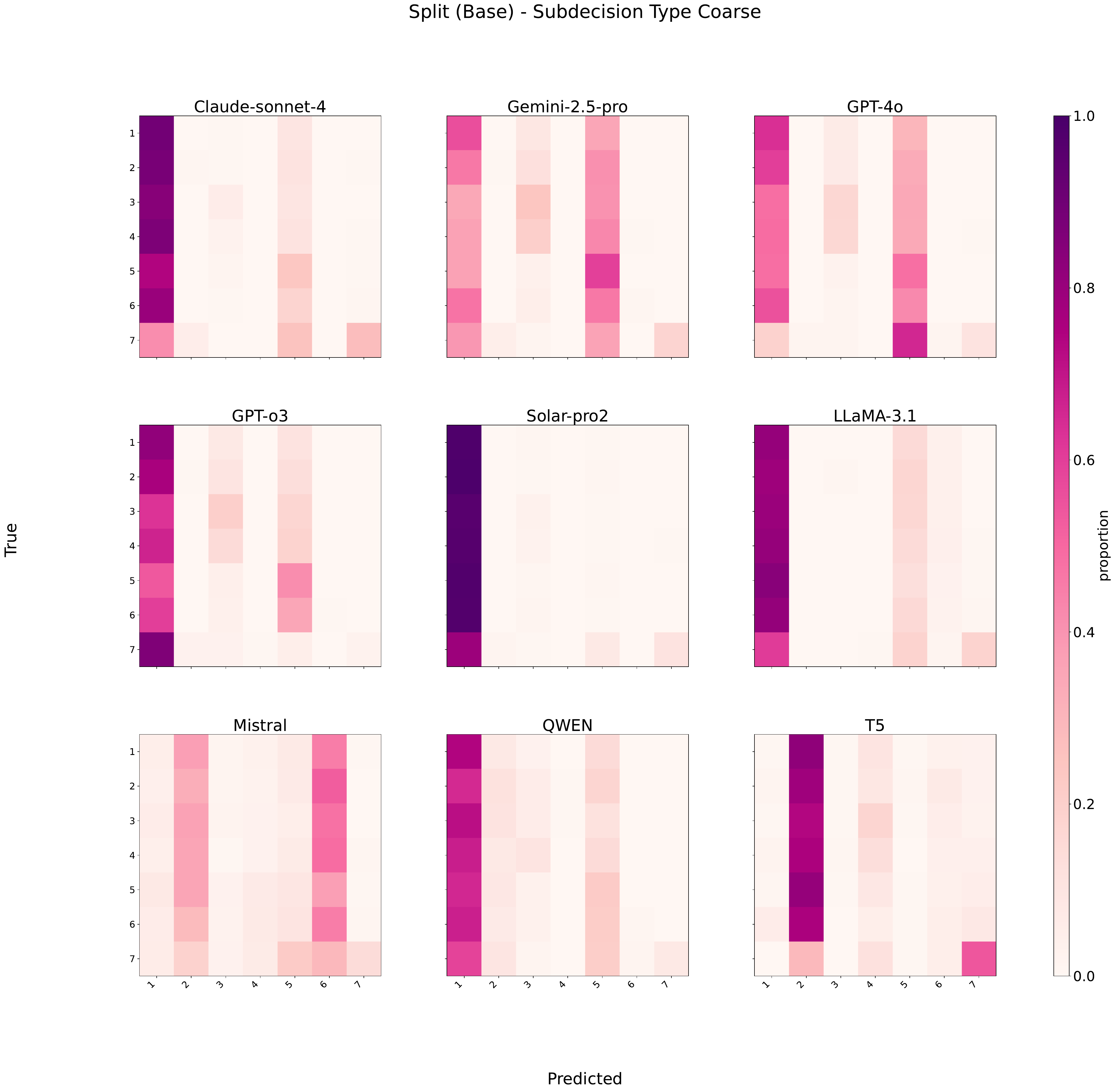}
    \caption{Heatmaps of model performance on the Subdecision (Coarse-grained) classification task under the Split (Base) input setting. Each subplot visualizes the distribution of predicted versus true labels across models. The numerical indices on the axes correspond to the canonical labels defined in Table~\ref{tab:subdecision-coarse-mapping-main}, where each index maps to a specific subdecision category.}
    \label{fig:base_subdecision_coarse_all_heatmap}
\end{figure*}

%  >>>>>>>

\begin{figure*}[t]
    \centering
    \includegraphics[width=\textwidth]{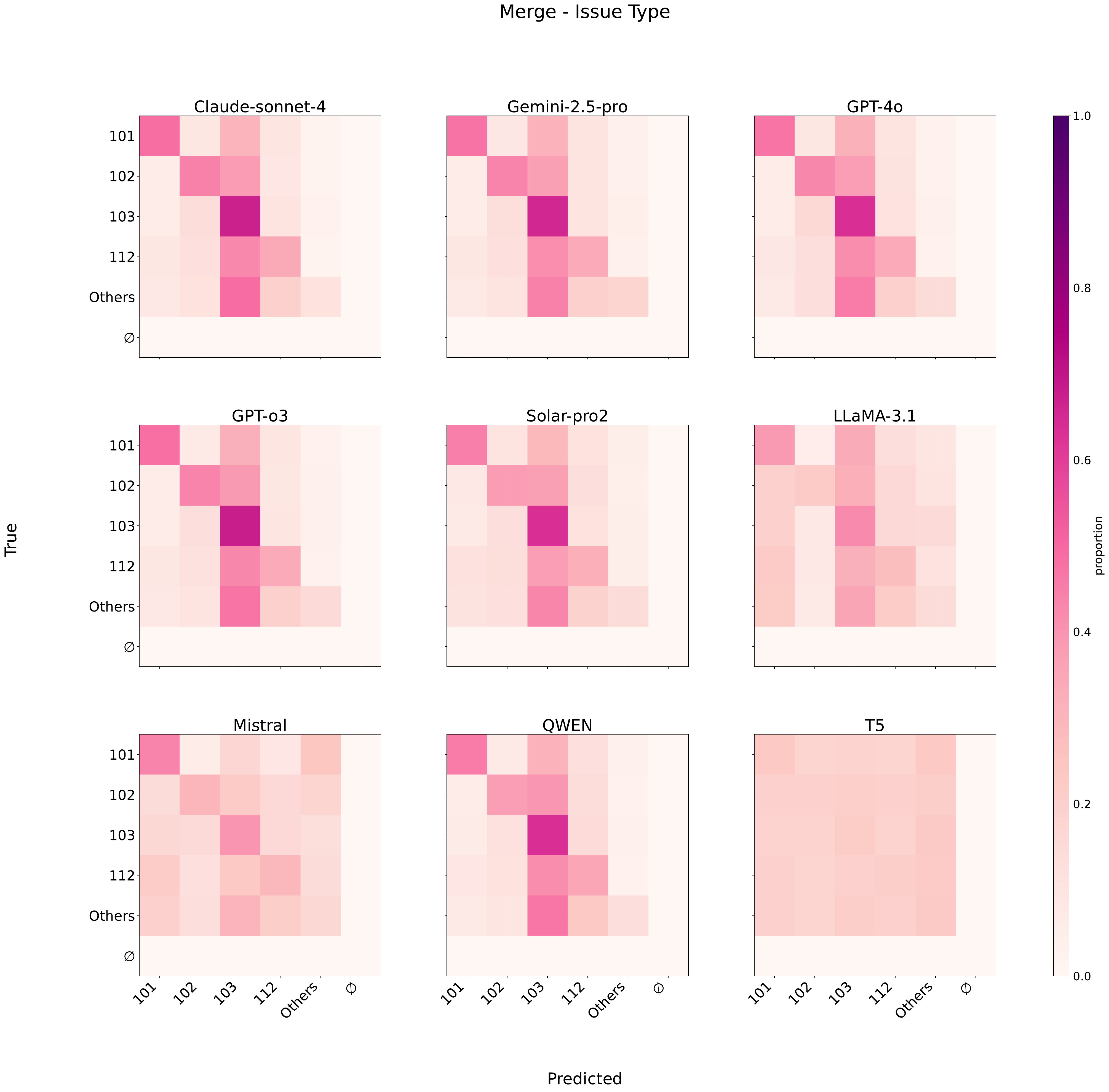}
    \caption{Heatmaps of model performance on the Issue Type classification task under the Merge input setting. Each subplot visualizes the distribution of predicted versus true labels across models.}
    \label{fig:merge_issue_type_all_heatmap}
\end{figure*}

\begin{figure*}[t]
    \centering
    \includegraphics[width=\textwidth]{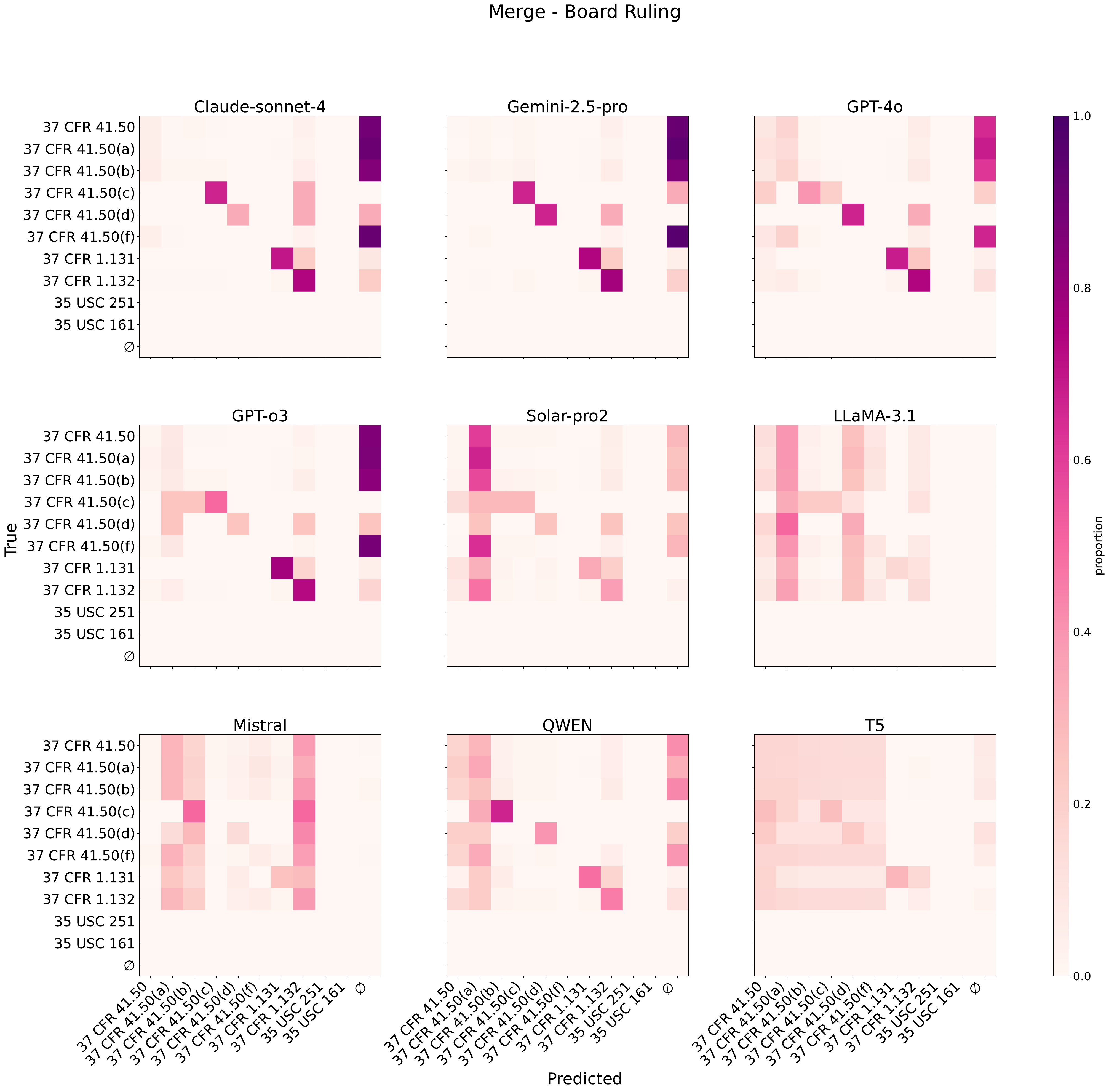}
    \caption{Heatmaps of model performance on the Board Authorities classification task under the Merge input setting. Each subplot visualizes the distribution of predicted versus true labels across models.}
    \label{fig:merge_board_authorities_all_heatmap}
\end{figure*}

\begin{figure*}[t]
    \centering
    \includegraphics[width=\textwidth]{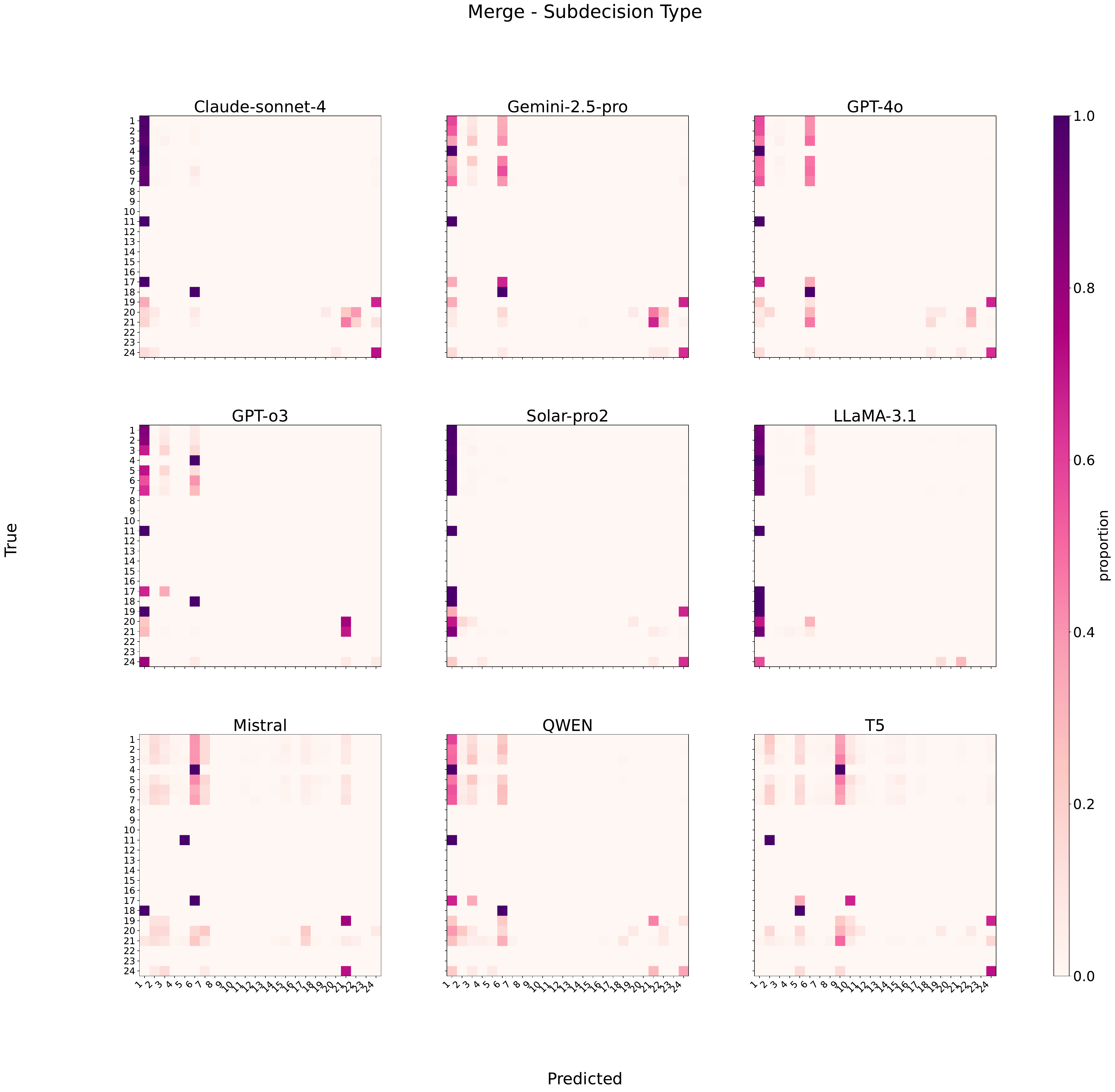}
    \caption{Heatmaps of model performance on the Subdecision (Fine-grained) classification task under the Merge input setting. Each subplot visualizes the distribution of predicted versus true labels across models. The numerical indices on the axes correspond to the canonical labels defined in Table~\ref{tab:subdecision-fine-mapping-main}, where each index maps to a specific subdecision category.}
    \label{fig:merge_subdecision_fine_all_heatmap}
\end{figure*}

\begin{figure*}[t]
    \centering
    \includegraphics[width=\textwidth]{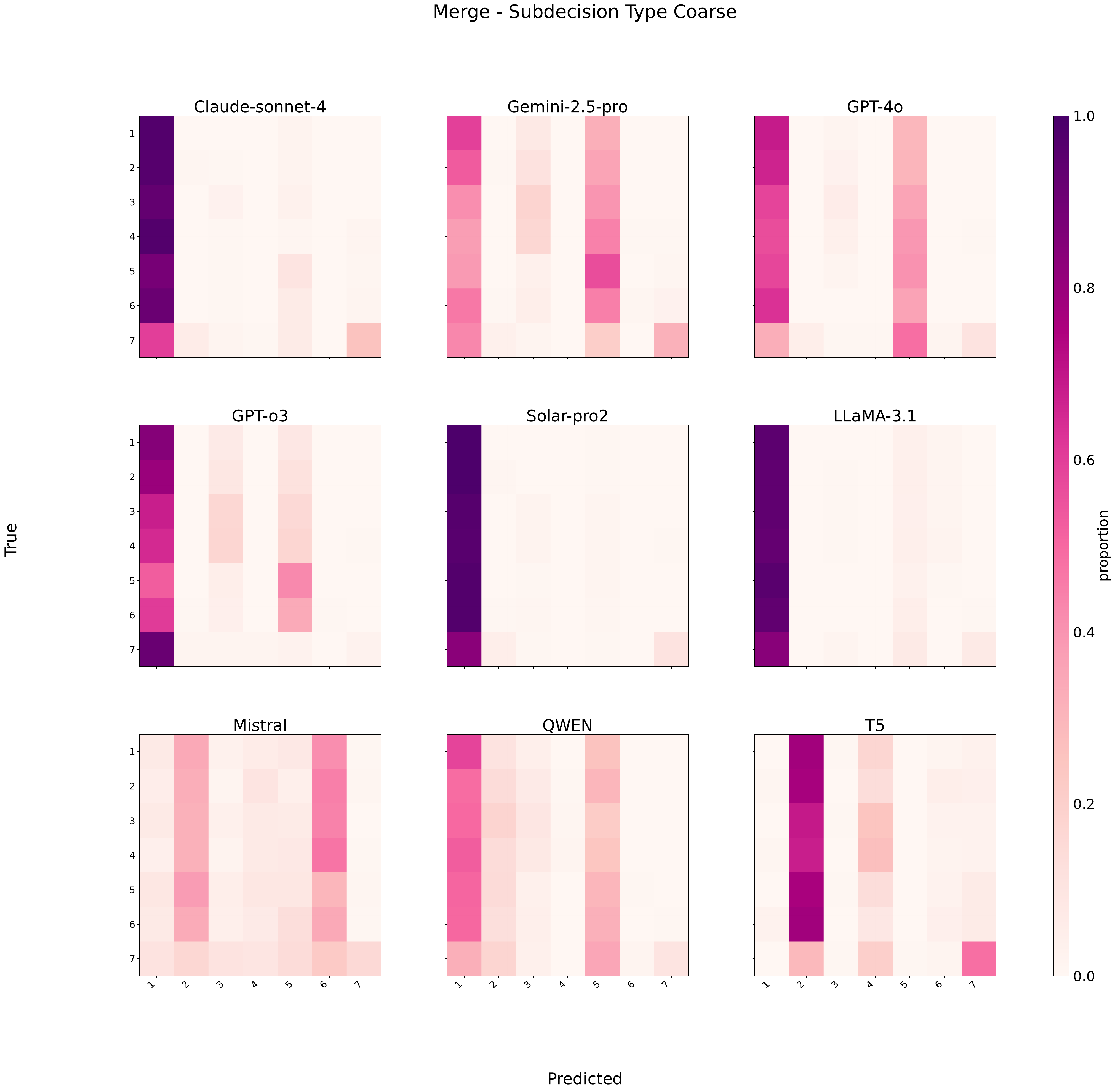}
    \caption{Heatmaps of model performance on the Subdecision (Coarse-grained) classification task under the Merge input setting. Each subplot visualizes the distribution of predicted versus true labels across models. The numerical indices on the axes correspond to the canonical labels defined in Table~\ref{tab:subdecision-coarse-mapping-main}, where each index maps to a specific subdecision category.}
    \label{fig:merge_subdecision_coarse_all_heatmap}
\end{figure*}

%  >>>>>>>

\begin{figure*}[t]
    \centering
    \includegraphics[width=\textwidth]{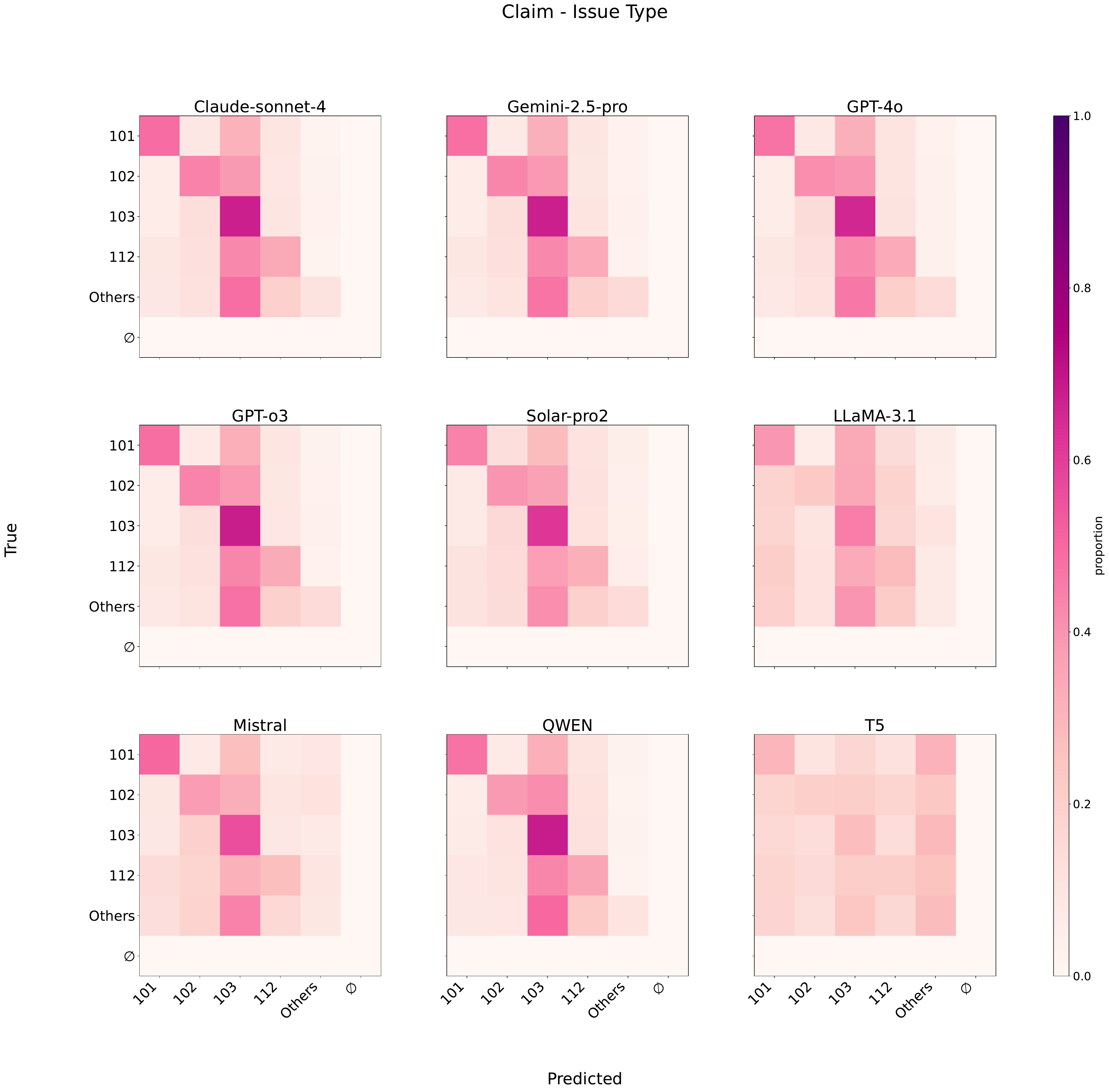}
    \caption{Heatmaps of model performance on the Issue Type classification task under the Split+Claim input setting. Each subplot visualizes the distribution of predicted versus true labels across models.}
    \label{fig:claim_issue_type_all_heatmap}
\end{figure*}

\begin{figure*}[t]
    \centering
    \includegraphics[width=\textwidth]{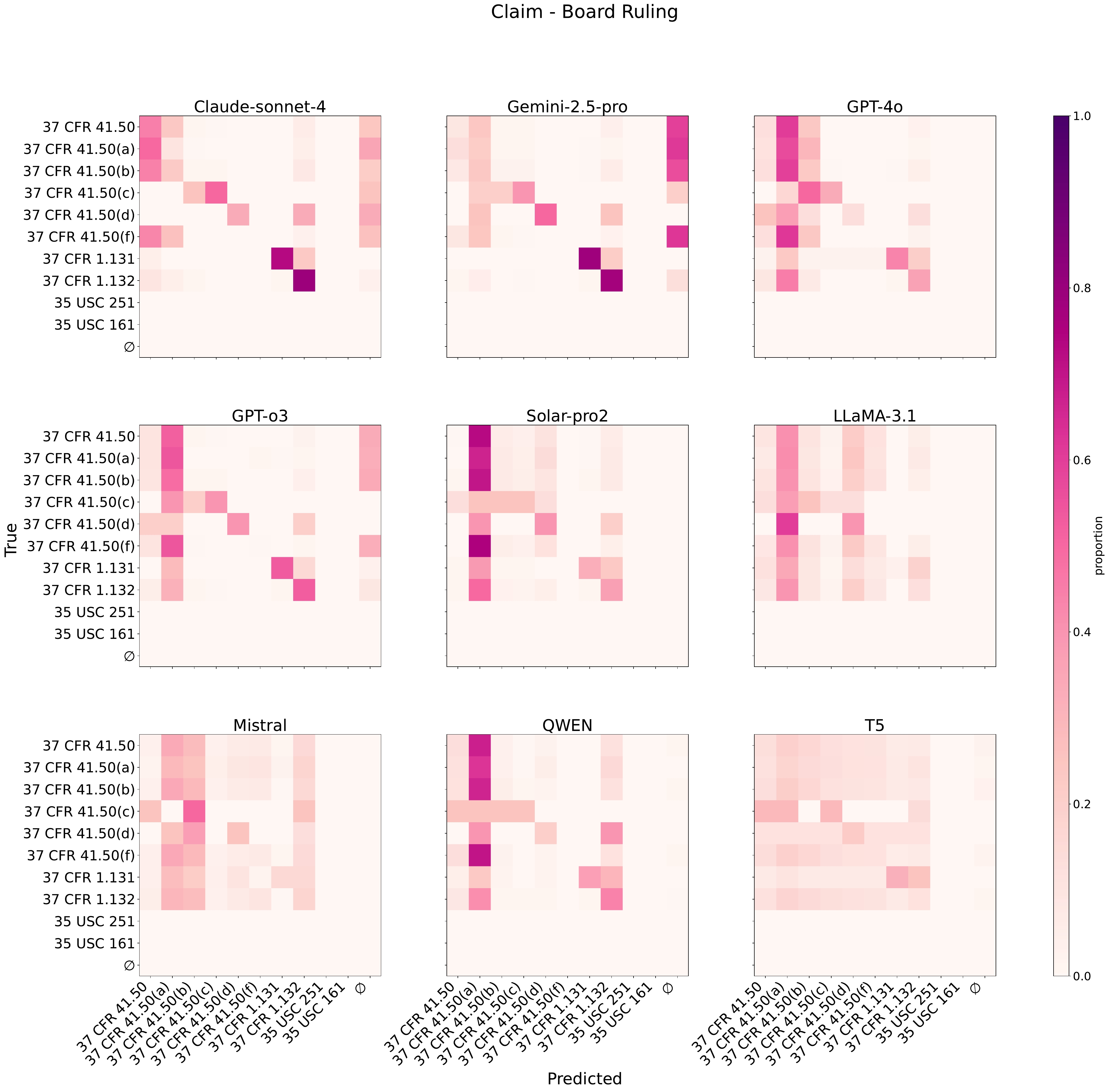}
    \caption{Heatmaps of model performance on the Board Authorities classification task under the Split+Claim input setting. Each subplot visualizes the distribution of predicted versus true labels across models.}
    \label{fig:claim_board_authorities_all_heatmap}
\end{figure*}

\begin{figure*}[t]
    \centering
    \includegraphics[width=\textwidth]{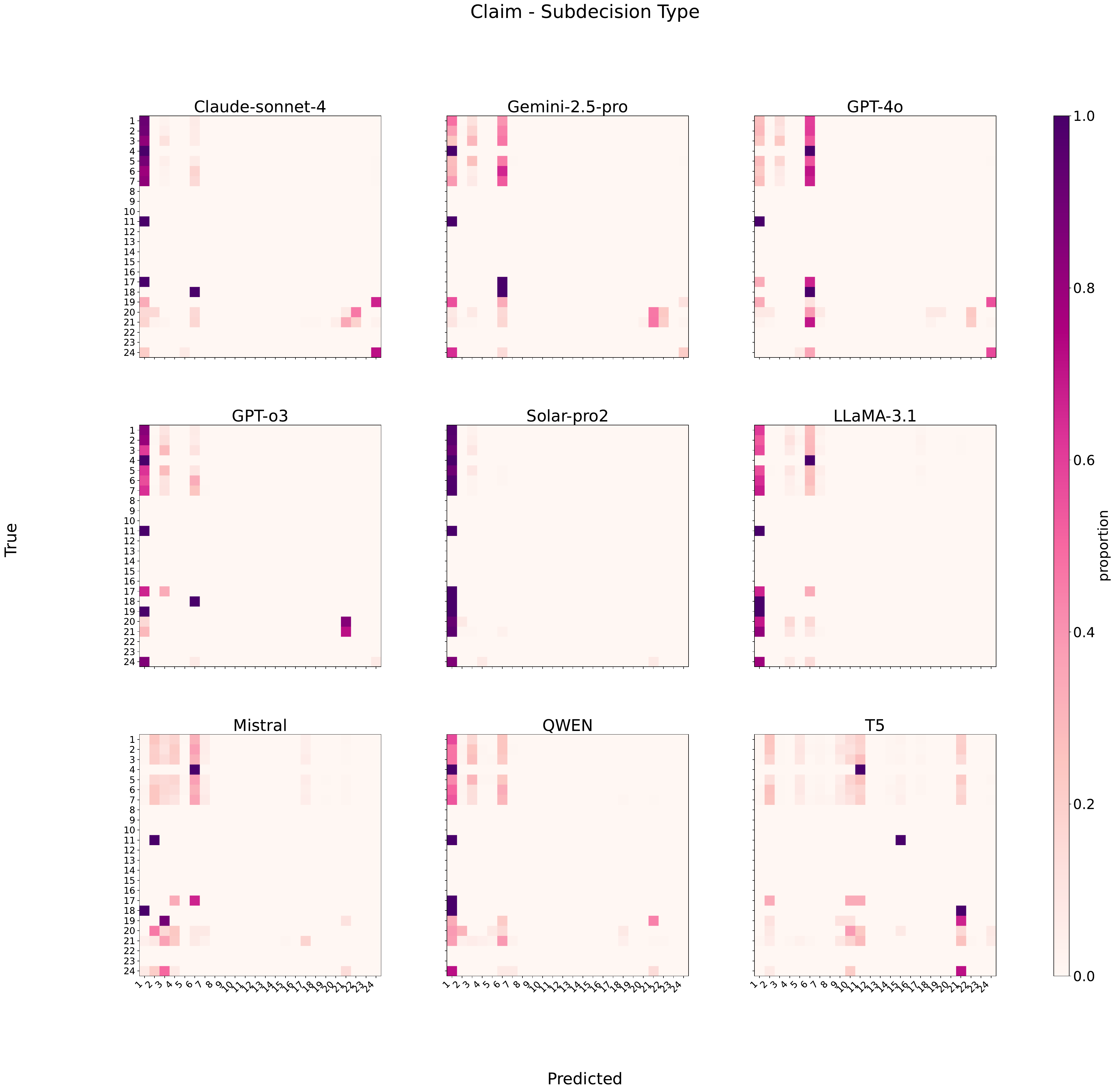}
    \caption{Heatmaps of model performance on the Subdecision (Fine-grained) classification task under the Split+Claim input setting. Each subplot visualizes the distribution of predicted versus true labels across models. The numerical indices on the axes correspond to the canonical labels defined in Table~\ref{tab:subdecision-fine-mapping-main}, where each index maps to a specific subdecision category.}
    \label{fig:base_subdecision_fine_all_heatmap}
\end{figure*}

\begin{figure*}[t]
    \centering
    \includegraphics[width=\textwidth]{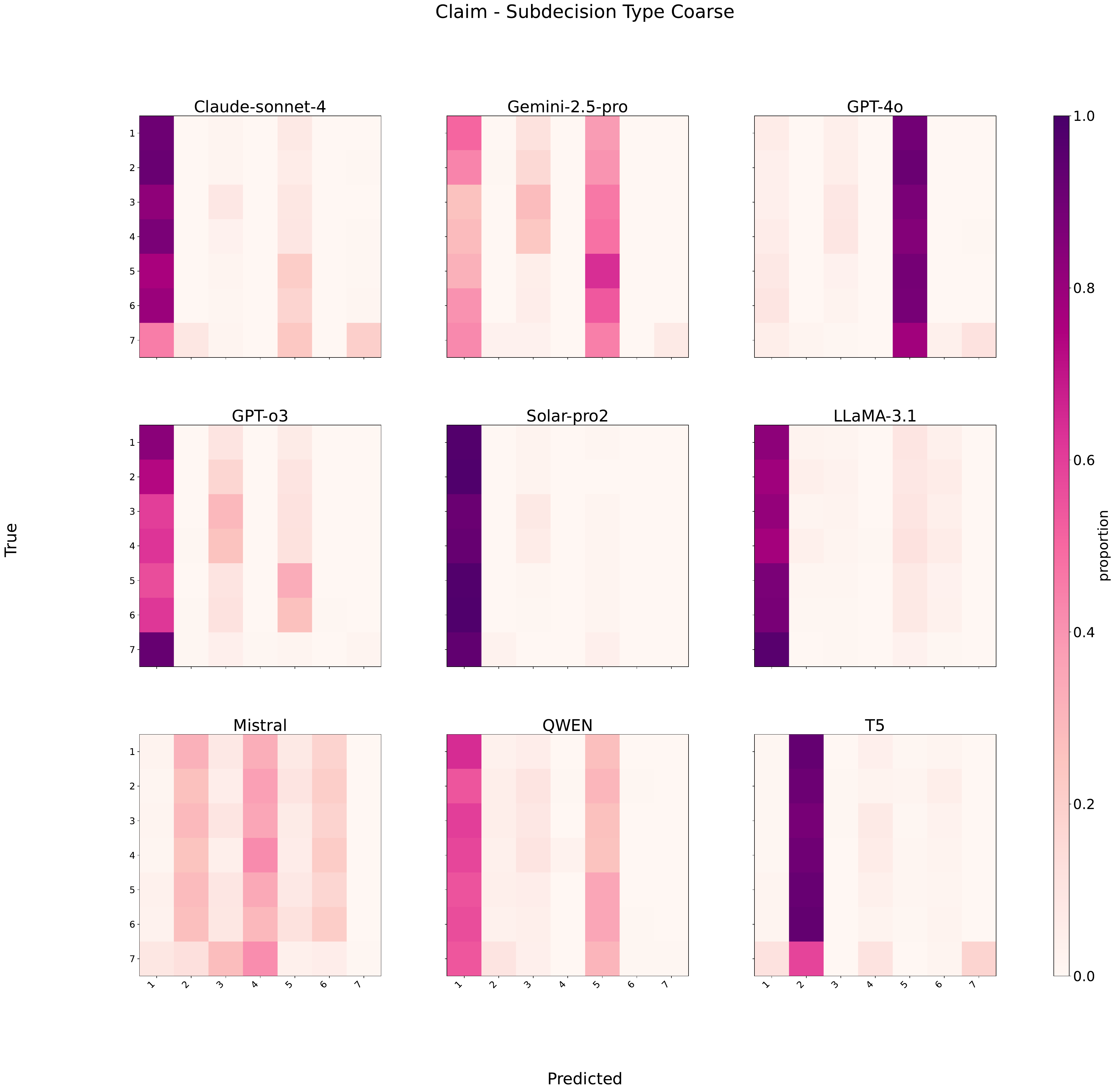}
    \caption{Heatmaps of model performance on the Subdecision (Coarse-grained) classification task under the Split+Claim input setting. Each subplot visualizes the distribution of predicted versus true labels across models. The numerical indices on the axes correspond to the canonical labels defined in Table~\ref{tab:subdecision-coarse-mapping-main}, where each index maps to a specific subdecision category.}
    \label{fig:claim_subdecision_coarse_all_heatmap}
\end{figure*}

\end{document}